\pdfoutput=1
\documentclass[11pt]{article}

\usepackage[]{acl}

\usepackage{times}
\usepackage{latexsym}

\usepackage[T1]{fontenc}
\usepackage[utf8]{inputenc}

\usepackage{microtype}
\usepackage{mathtools}  
\usepackage{subcaption}
\usepackage{xcolor, colortbl}
\usepackage{booktabs}
\usepackage{algorithm}
\usepackage{algorithmic}
\usepackage[utf8]{inputenc}
\usepackage{epsfig}
\usepackage{graphicx}
\usepackage{amssymb}
\usepackage{epsfig}
\usepackage{amsmath, mdframed}
\usepackage{multirow}
\usepackage{makecell}
\DeclareMathOperator*{\argmax}{arg\,max}
\newcolumntype{P}[1]{>{\centering\arraybackslash}p{#1}}
\usepackage{color,soulutf8}
\usepackage[export]{adjustbox}

\newcommand{\DITTO}{\textit{DiTTO}\includegraphics[height=3ex,valign=m]{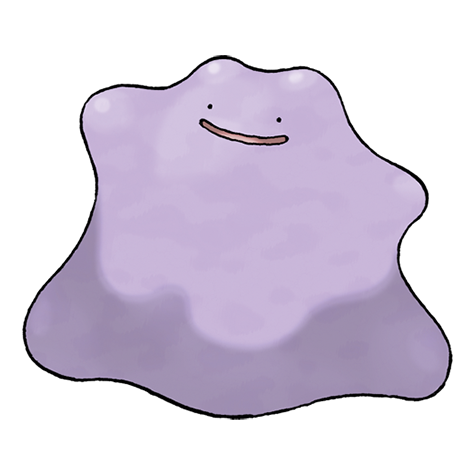} }

\newcommand{\DiTTO}{\textit{DiTTO} }

\title{\DITTO: A Feature Representation Imitation Approach for Improving Cross-Lingual Transfer}

\author{ Shanu Kumar \quad Abbaraju Soujanya \quad Sandipan Dandapat\\
\textbf{ Sunayana Sitaram  \quad Monojit Choudhury}\\
Microsoft Corporation, India \\
{\tt \small \{shankum,asoujanya,sadandap,susitara,monojitc\}@microsoft.com} }

\begin{document}
\maketitle
\begin{abstract}
Zero-shot cross-lingual transfer is promising,  however has been shown to be sub-optimal, with inferior transfer performance across low-resource languages. In this work, we envision languages as domains for improving zero-shot transfer by jointly reducing the feature incongruity between the source and the target language and increasing the generalization capabilities of pre-trained multilingual transformers. We show that our approach, \DiTTO, significantly outperforms the standard zero-shot fine-tuning method on multiple datasets across all languages using solely unlabeled instances in the target language. Empirical results show that jointly reducing feature incongruity for multiple target languages is vital for successful cross-lingual transfer. Moreover, our model enables better cross-lingual transfer than standard fine-tuning methods, even in the few-shot setting.
% Our observations show that this sub-optimality is due to incongruency in feature representations of source and target languages, resulting in poor generalization of these over-parameterized MMTs.
% This leads us to a formulation that encapsulates both generalization capability and feature incongruity to improve the cross-lingual transfer of any MMT

% We further show that improvements are not limited to high-resource languages, but it generalizes well to resource-lean, distant, and unseen languages with more significant gains. 
% We also evaluate \textit{DiTTO} in terms of cost and find it seven times cheaper than existing solutions for this problem. 
\end{abstract}

\section{Introduction}

Due to the emergence of pre-trained Massively Multilingual Transformers (MMTs) such as mBERT \cite{devlin-etal-2019-bert}, XLM-R \cite{conneau2020unsupervised} and mT5 \cite{xue2020mt5}, zero-shot cross-lingual transfer \cite{hu2020xtreme, ruder-etal-2021-xtreme, lauscher2020zero, ansell2021mad, pfeiffer-etal-2022-lifting} has received  significant attention in the NLP community. This approach originated due to the skew in resource distribution in languages \cite{joshi-etal-2020-state}, with most languages of the world having a scarcity of labeled data. 
% The inception of this approach originates from the fact that resource distribution across languages is skewed towards English and few other languages \cite{joshi-etal-2020-state}, leaving the rest of them with the scarcity of labeled data. 
% Zero-shot involves initially fine-tuning the MMT with the task-specific data in one or multiple source languages and is evaluated upon a target language, none of whose labeled instances were used during fine-tuning. Hence, low-resource languages are crucially dependent on zero-shot cross-lingual transfer for specific tasks.
 Zero-shot transfer involves fine-tuning the MMT with task-specific data in one or more source languages, followed by evaluation on target languages whose labeled instances are not used during fine-tuning. Accurate zero-shot transfer is crucially important for MMTs to be useful for low-resource languages.
% \begin{figure}[t]
% \minipage{0.23\textwidth}
%   \includegraphics[width=\linewidth]{figures/baseline_3d_language_v2.png}
%   \caption*{without DiTTO}
% \endminipage
% \hfill\minipage{0.23\textwidth}
%   \includegraphics[width=\linewidth]{figures/ditto_3d_language_v2.png}
%   \vspace{-1em}
%   \caption*{with DiTTO}
% \endminipage
% \vspace{-0.7em}
%   \caption{The figure shows the 3D t-SNE visualization of the features from the last layer of the MMT (mBERT) after fine-tuning on 1\% of the source language data. Detailed visualizations in Figures \ref{fig:baseline_3dtsne} and \ref{fig:ditto_3dtsne} of Appendix.}
%     \label{fig:3d_tsne}
%     \vspace{-1em}
% \end{figure}

The performance of MMTs drops in the following two cases - when the source and target languages exhibit dissimilar typological features, or  when the size of pre-training data in the target language is limited \cite{lauscher2020zero, ebrahimi-etal-2022-americasnli}.
% Though MMTs have shown impressive zero-shot cross-lingual abilities \cite{pires-etal-2019-multilingual, wu-dredze-2019-beto, hu2020xtreme, ruder-etal-2021-xtreme}, their performance seems to drop when either typological dissimilarity exists between the source and target language, or when the size of labeled data from target language is very limited in the pre-training process of MMT \cite{lauscher2020zero, ebrahimi-etal-2022-americasnli}. 
Two common techniques to improve zero-shot performance include few-shot cross-lingual transfer \cite{lauscher2020zero, kumar-etal-2022-diversity} and the translate-train approach \cite{ruder-etal-2021-xtreme, ahuja-etal-2022-economics}. Several studies have been conducted comparing these approaches, of which \cite{ahuja-etal-2022-economics} concludes that if the cost of machine translation is greater than zero, the optimal and lowest-cost performance is achieved with at least some manually labeled data (i.e. the few-shot method). Since annotating data is expensive for many languages \cite{dandapat-etal-2009-complex, 10.1145/2362456.2362479, fort2016collaborative}, we investigate improving cross-lingual zero-shot transfer using only unlabelled data in this paper.

% There are also studies that suggest that annotating data for many languages is expensive and time-consuming \cite{dandapat-etal-2009-complex, 10.1145/2362456.2362479, fort2016collaborative}, however getting unlabeled examples is relatively very cheap and can be scaled for many languages.
% All these solutions are banking upon infusing the fine-tuning data with data of low-resource target language or manipulating source data to compensate for the missing data from the target language. But why do we need to invest more effort or cost to improve zero-shot, is there a better way? We deep-dived into zero-shot to search for an answer to this question.

% \begin{figure}[t]
% \centering
% \includegraphics[scale=0.15]{figures/baseline_ru_mbert_ADAM_1_0_15_14_Lang-Label.png}
% \caption{The plot shows 3D t-SNE visualization of the features from the last layer of the mBERT learnt using \textit{Baseline} method in XNLI-1 dataset. The features of the source language are denoted by ({\color{red} {\small EN}}) and the target language as ({\color{blue} {\small RU}}).
% }
% \label{fig:baseline_3d}
% \end{figure}

% Reason is, Zero-shot learning is unstable
Zero-shot Cross-lingual Transfer has been identified as an under-specified optimization problem \cite{wu2022zero}. A majority of the solutions reports a high performance on the source language but fluctuating performance on target languages. \citet{wu2022zero} use linear interpolation to prove that it is possible to obtain a subset of solutions which have optimal performance on both source and target languages.
% Zero-Shot Cross-lingual Transfer has been identified as an under-specified optimization problem \citet{wu2022zero} where the task-specific labeled data from only the source language is used to optimize an MMT, leading to many solutions having high performance on the source language but fluctuating performance on the target language. The mentioned study also uses linear interpolation to prove that there is a possibility that we can reach a subset of solutions that has optimal performance on both source and target language. 
Furthermore, they also conclude that current optimization techniques cannot converge to this smaller subset of optimal solutions without the availability of labeled target language data. \citet{aghajanyan2020better} and \citet{liu-etal-2021-preserving} have observed similar behavior in the zero-shot setup and hypothesize that sub-optimal zero-shot performance may be due to the degradation of generalizable representations of MMTs during the fine-tuning stage. This leads to the model trained on the source language not being able to generalize well to the target languages. MMTs have also been shown to be over-parameterized \cite{l.2018a, kolesnikov2020big, zhang2021understanding}, which leads to memorizing the training data (source language) and achieving poor generalization during cross-lingual transfer.
% They further conclude that current optimization techniques could not converge to this smaller subset without the availability of labeled target language data. Few works \cite{aghajanyan2020better, liu-etal-2021-preserving} have observed similar behavior in the zero-shot setup. They theorize that the sub-optimal zero-shot performance can be due to the degradation of generalizable representations of MMTs during the fine-tuning stage. Due to this, the model learned on the source language will not be generalized to the target language. Since MMTs are over-parameterized \cite{kolesnikov2020big, zhang2021understanding}, they can quickly memorize the training data (source language) and achieve poor generalization, which snowballs into an inferior cross-lingual transfer.

%There are a lot of good solutions but the common losses which we use have a lot of local minima, and because loss landscapes are non-convex, various minimas are emerging leading to fluctuating performances.
% Our hypothesis of languages being domains and SAM

\begin{figure}[t]
\centering
\includegraphics[scale=0.35]{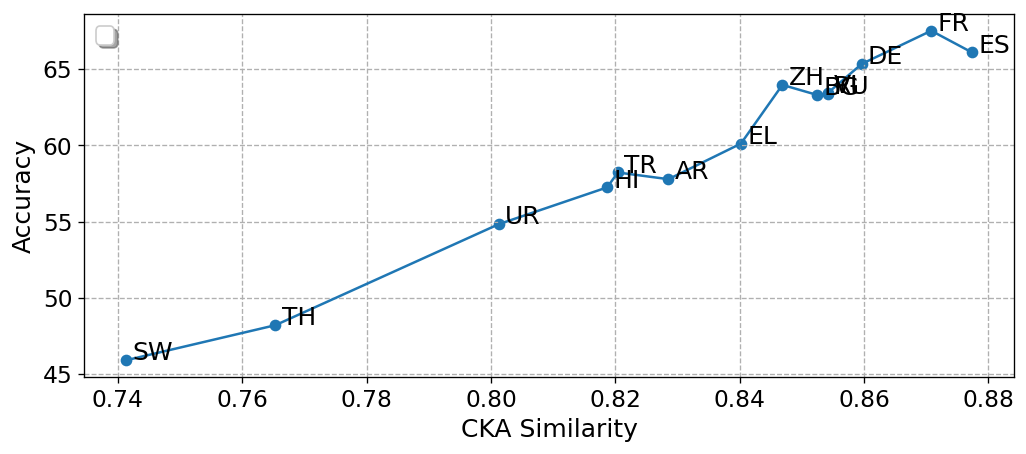}
\vspace{-1em}
\caption{Relation between the zero-shot performance using mBERT, and CKA similarity between the source ({\small EN}) and various target languages in XNLI dataset. }
\label{fig:baseline_cka}
\vspace{-1em}
\end{figure}

 Similar to \citet{deshpande-etal-2022-bert}, in our experiments, we also observe that once MMTs are fine-tuned on source languages, there is an incongruity between the features of the source and target languages, as shown in Figure \ref{fig:ditto_3dtsne}. We speculate that the mismatch in the feature representation space causes problems in generalization. We also find that this mismatch strongly correlates with zero-shot performance as shown in Figure \ref{fig:baseline_cka}.
 
 Furthermore, we hypothesize that this instability can be reduced either by finding solutions that can generalize well or learning to match the feature representations. Solutions \cite{zhang2018mixup, Jiang*2020Fantastic} that have been used for improving generalization in other tasks can be considered, so that the model reaches to a better local minima. Sharpness-aware Optimization (SAM) \cite{foret2021sharpnessaware} is one such technique that has been used to improve the generalization of language models \cite{bahri2022sharpness} and vision transformers \cite{chen2021vision} by smoothing the loss landscape for various adversarial tasks. SAM is used to generalize across domains, however, by treating languages as separate domains, we can apply SAM for generalizing across languages. While SAM looks promising, our experiments (cf. \ref{section:ablation}) showed that it does not guarantee optimal generalization at all times. We need to further reduce the incongruency between language features by aligning target language features to mimic the features of the source language. We propose \textbf{\DITTO} for improving cross-lingual transfer by source language \textbf{Di}rected adversarial \textbf{T}ransition of \textbf{T}arget language using sharpness aware \textbf{O}ptimization.

The key contributions of this work are: 
1) Exhibiting the limitations of standard fine-tuning by unveiling the feature incongruity between source and target languages.
%2) As we envisioned languages as domains, it helped in curation of a method \DiTTO to enhance cross-lingual transfer by feature transformation of the target language to mimic the source.
2) \DiTTO enhances cross-lingual transfer by joint feature transformation of the multiple target languages to mimic the source.
3) \DiTTO makes cross-lingual transfer cost-effective and efficient for distant (typologically different languages), resource-lean and unseen (not present in the pre-training data) languages.
4)\DiTTO exhibits superior performance compared to augmenting
% \footnote{augmenting refers to increasing the size of labeled data used during fine-tuning}
the training data for either the source or the target language.

% [1] We show the limitations of the standard fine-tuning method using incongruity between the features of the source and target language.
% [3] Through experiments on multiple datasets, we show improvements using \textit{DiTTO} over the standard training method in both zero-shot and few-shot settings.
% We illustrate the performance of our method on various dimension: i) improvements on datasets consisting of languages both seen and unseen during pre-training, 
% ii) cost, and iii) improves congruity between the features representations of source and target languages.

% Contribution of the work:
% \begin{itemize}
%     \item DiTTO improves zero-shot pr without labeled data
%     \item improves Calibration 
%     \item EVEN IMPROVES FOR UNSEEN LANGUAGE
%     \item we are evaluating the proposed method on 4 text classification tasks.
% \end{itemize}

\section{Related Work}
\textbf{Cross-Lingual Transfer}: Since the inception of pre-trained MMTs, zero-shot learning has become popular for cross-lingual tasks. Recent works \cite{lauscher2020zero,  ebrahimi-etal-2022-americasnli, wu2022zero} have shown it to be sub-optimal for target languages which are either distant to the source language or have limited data during pre-training of the MMT.
Some works \cite{wu-dredze-2020-explicit, yu2021effective} have tried to improve the transfer using feature alignment from parallel data or bitexts \cite{zhang2020improving, tiedemann-2012-parallel} which is often expensive to obtain for many languages. To address this issue, \DiTTO relies only on unlabeled data in the target languages.
As pre-training size of the language affect transfer performance, adapter-based frameworks \cite{pfeiffer-etal-2020-mad, ansell2021mad} have been proposed for learning language and task representations for low-resource languages and languages that are unseen during pre-training. Though this framework is helpful for unseen languages, it provides limited gains for typologically dissimilar and high resource languages, and our method can easily be integrated with adaptors to further improve the transfer performance. 

\hspace{-1em}\textbf{Improving Generalization}: Deep neural networks such as MMTs are generally over-parameterized and fine-tuning leads to easy memorization of the labeled training data, does not always generalize well to other domains \cite{l.2018a, kolesnikov2020big, zhang2021understanding}. Various methods have been proposed to improve the generalization like dropout \cite{JMLR:v15:srivastava14a}, label smoothing \cite{muller2019does}, batch normalization \cite{ioffe2015batch}, mixup \cite{zhang2018mixup}.
% , and data augmentation \cite{cubuk2018autoaugment}. 
% \citet{keskar2017on} have shown that large-batch tend to converge towards a sharp minmia, leading to poor generalization. 

A few papers \cite{dziugaite2017computing, NEURIPS2019_01d8bae2, Jiang*2020Fantastic} have explored the connection between the flatness of minima and generalization gaps, showing flatter minima leads to better generalization. Recently, SAM has been proposed to find a smoother minima by minimizing the loss value and its sharpness. SAM has been shown to improve the generalization capabilities of vision transformers \cite{chen2021vision}. Recently, \citet{bahri2022sharpness} employed SAM in language models such as GPT-3 \cite{brown2020language} and T5 \cite{raffel2020exploring}, showing significant improvements in generalization in English. In this work, we use SAM to improve the generalization across other languages. Another line of work \cite{aghajanyan2020better, liu-etal-2021-preserving} hypothesizes that inferior transfer is due to forgetting and degradation of feature representation from pre-trained MMTs when they are fine-tuned on the source language data. They propose to preserve the pre-trained features to improve the generalization using regularization and continual learning.

\hspace{-1em}\textbf{Unsupervised Domain Adaptation (UDA)}: 
% It has been proposed to address the issue of models trained on the source domain, do not generalize well to the target domain with some distributional shift. 
Various studies have been proposed to reduce the domain shift to perform UDA by minimizing discrepancy distances such as Maximum Mean Discrepancy (MMD) \cite{long2015learning} and correlation alignment distance \cite{sun2016deep}. Adversarial-based feature alignment methods \cite{ganin2015unsupervised, ganin2016domain, long2018conditional, kurmi2019attending} have been one of the popular UDA methods where the domain discrepancy between the domains is reduced using an adversarial objective. In this work, we use Domain-Adversarial Neural Networks (DANN) \cite{ganin2016domain} for performing adversarial adaptation of languages. 
% \citet{kandula2021improving} has performed adversarial adaptation to improve performance on the target language but it is limited to only bilingual settings, whereas our approach tries to improve the performance in all the languages.

\section{Background}
\textbf{Training a Zero-Shot Model}:
In zero-shot cross-lingual transfer, we fine-tune an MMT on a source language and evaluate its performance on the target language, whose instances are not used during fine-tuning. To do this, we need a source language $s$ and task-specific labeled dataset $\mathbb{L}_s = \{(x^s_i, y^s_i)\}^{n}_{i=1}$ with $n$ examples. We use the provided MMT $\mathcal{M}$ as the encoder and fine-tune it along with the task-specific classifier $\mathcal{C}$ by minimizing the cross-entropy loss:
\vspace{-0.5em}
\begin{equation}
\label{eqn:standard_zeroshot}
\begin{aligned}
\mathcal{L}_\text{train} (\mathcal{M}, C) = \mathbb{E}_{(\textbf{x}^s_i, \textbf{y}^s_i) \sim \mathbb{L}_s}\mathcal{L}(C(\mathcal{M}(\textbf{x}^s_i)), \textbf{y}^s_i)   \\
\end{aligned}
\end{equation}
\vspace{-0.5em}

\begin{figure*}[!t]
\begin{subfigure}[t]{\textwidth}
    \minipage{0.31\textwidth}
  \includegraphics[width=\linewidth]{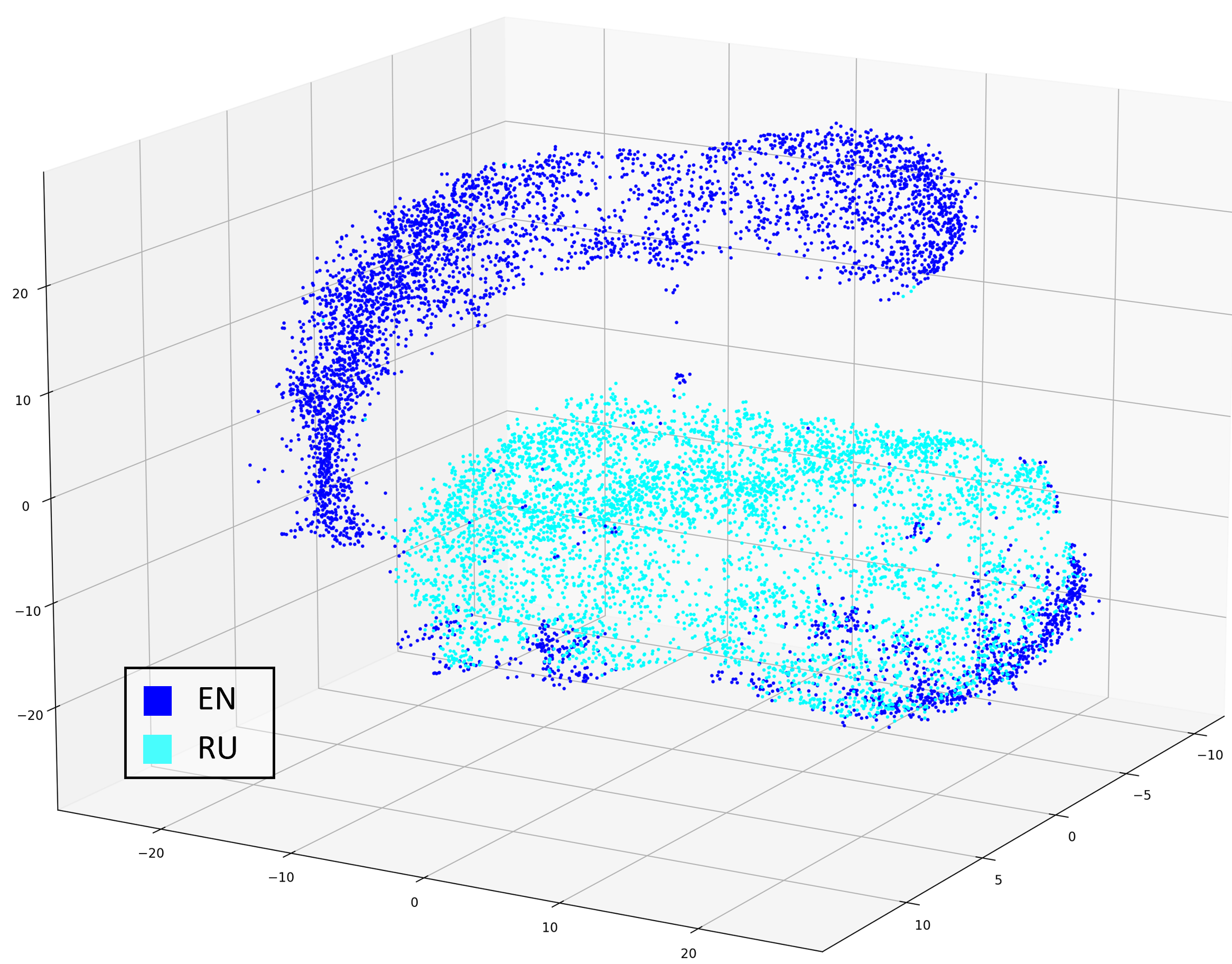}
  \vspace{-1em}
  \subcaption{Language-wise Labels (Zero-shot)}
    \label{fig:zeroshot_language}
\endminipage\hfill\minipage{0.31\textwidth}
  \includegraphics[width=\linewidth]{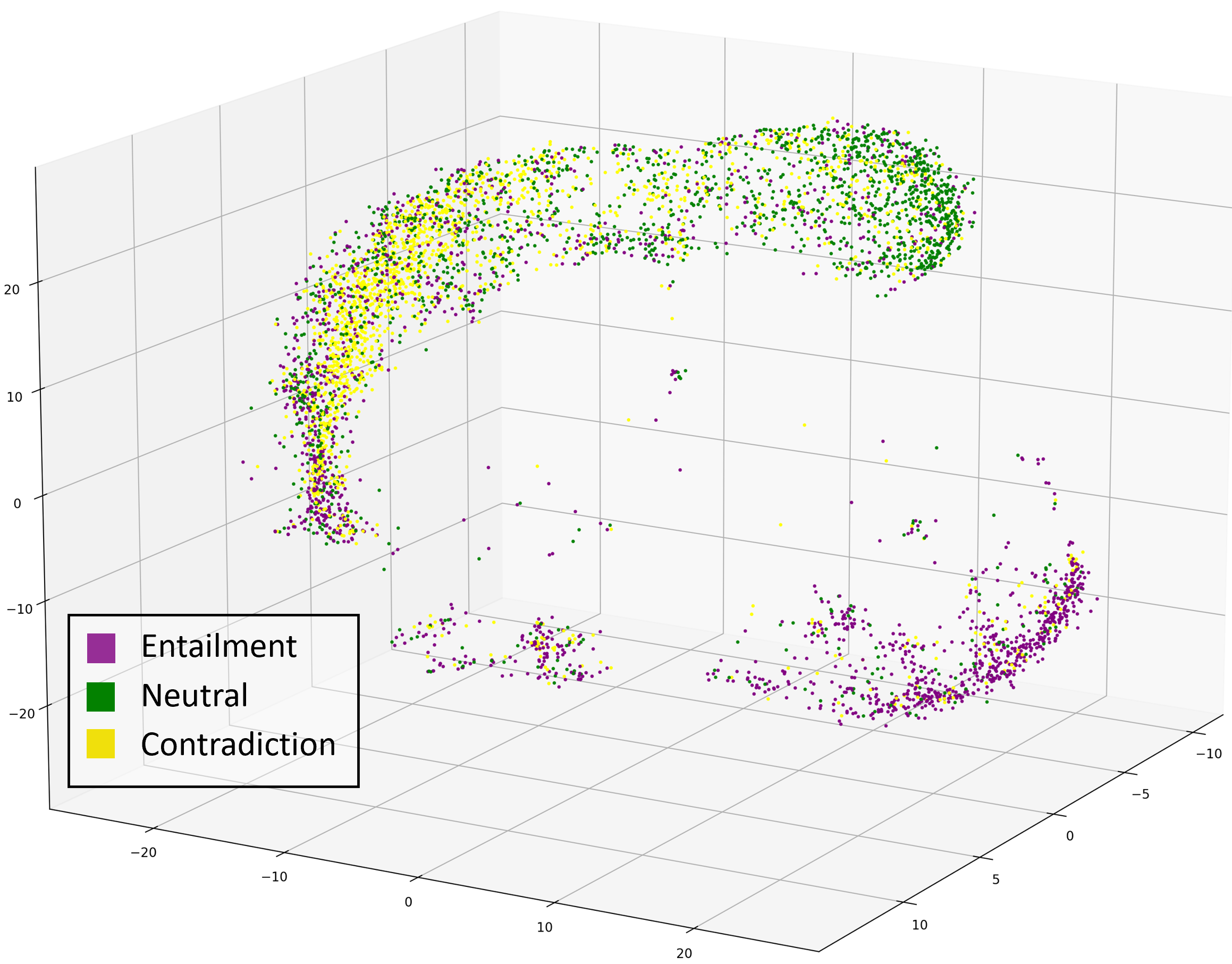}
  \vspace{-1em}
  \subcaption{Class-wise Labels (Source: {\small EN})}
    \label{fig:zeroshot_source}
\endminipage\hfill\minipage{0.31\textwidth} \includegraphics[width=\linewidth]{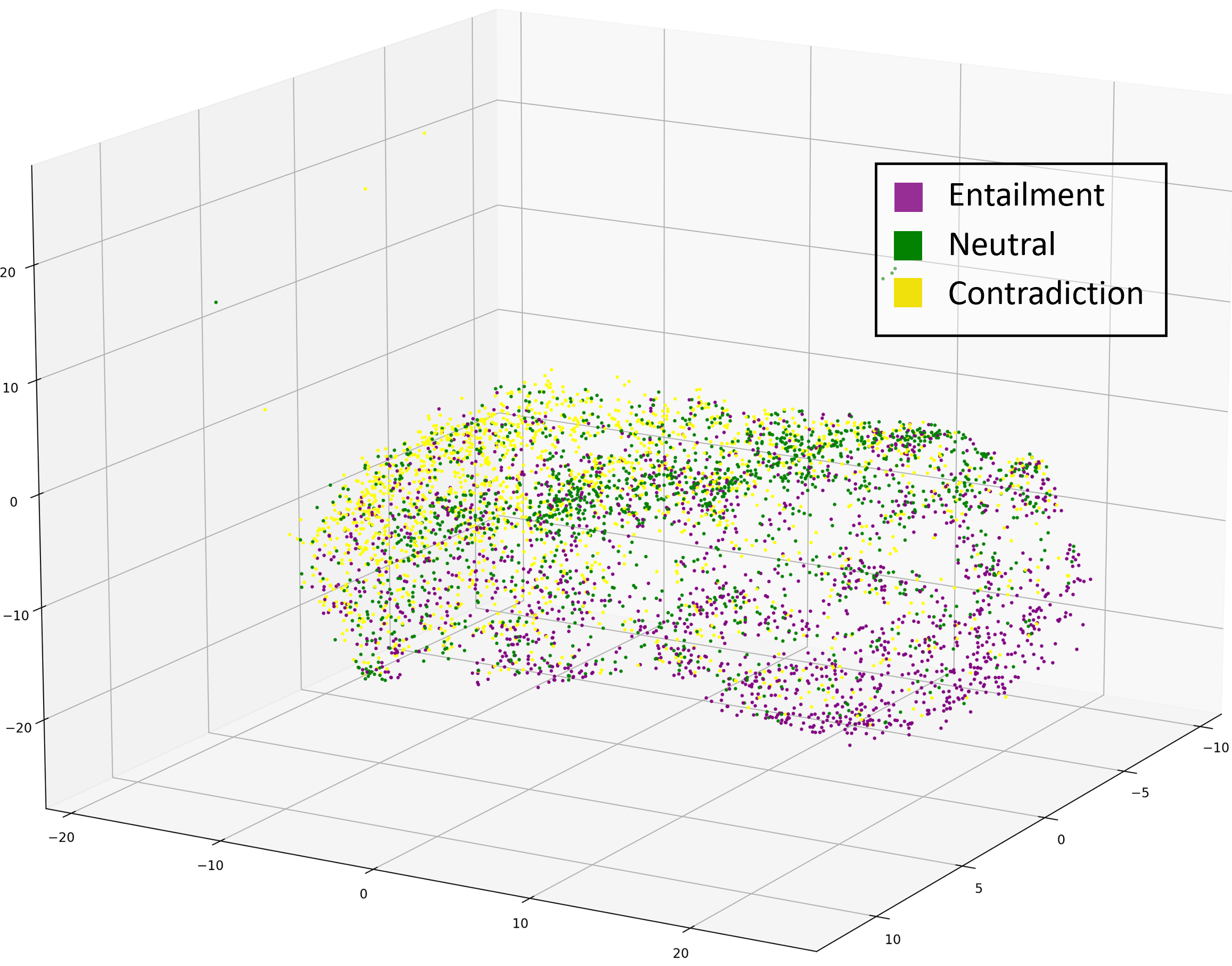}
\vspace{-1em}
  \subcaption{Class-wise Labels (Target: {\small RU})}
    \label{fig:zeroshot_target}
\endminipage
\end{subfigure}
\begin{subfigure}[t]{\textwidth}
\minipage{0.32\textwidth}
  \includegraphics[width=\linewidth]{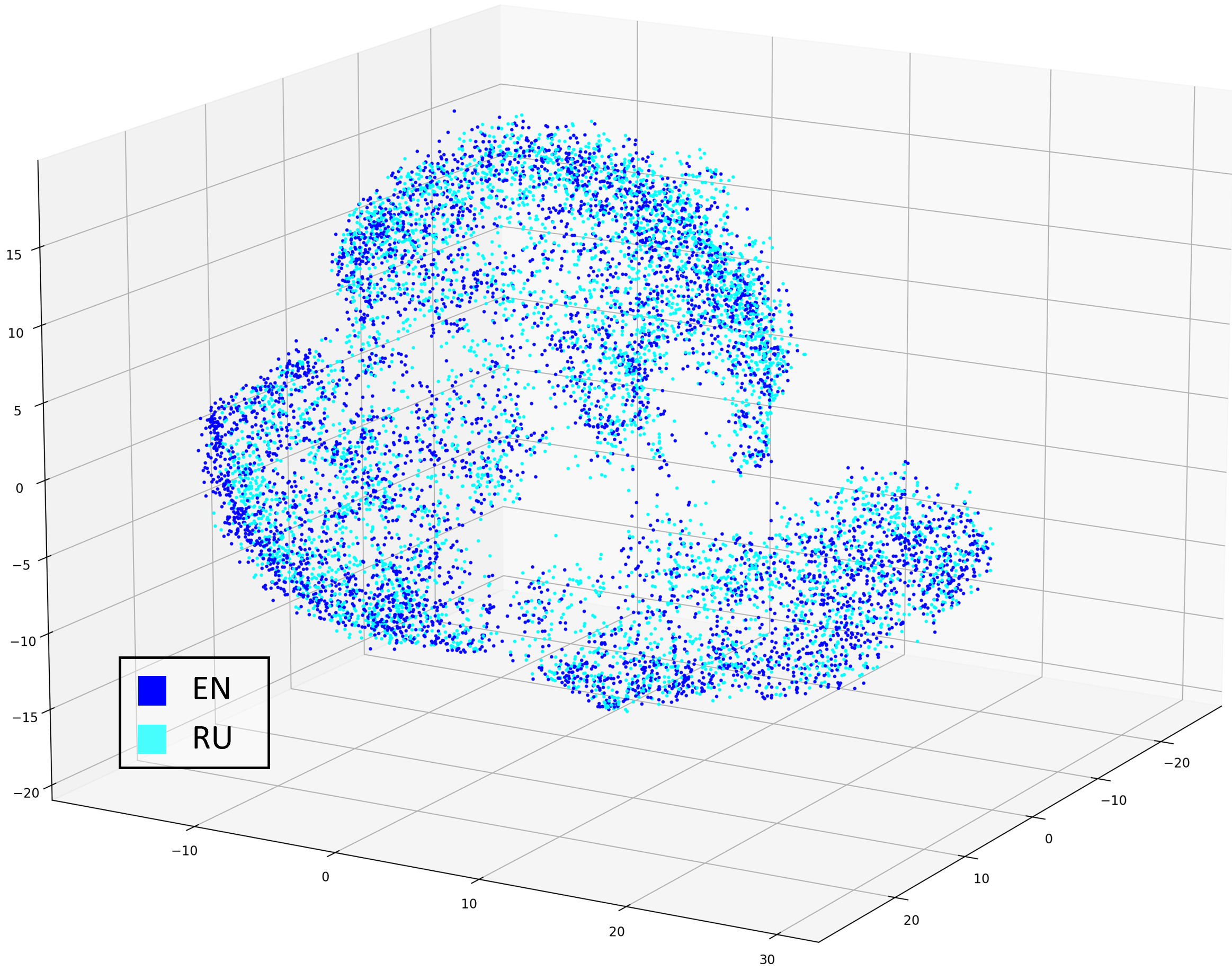}
  \subcaption{Language-wise Labels (\textit{DiTTO})}
\endminipage\hfill\minipage{0.31\textwidth}
  \includegraphics[width=\linewidth]{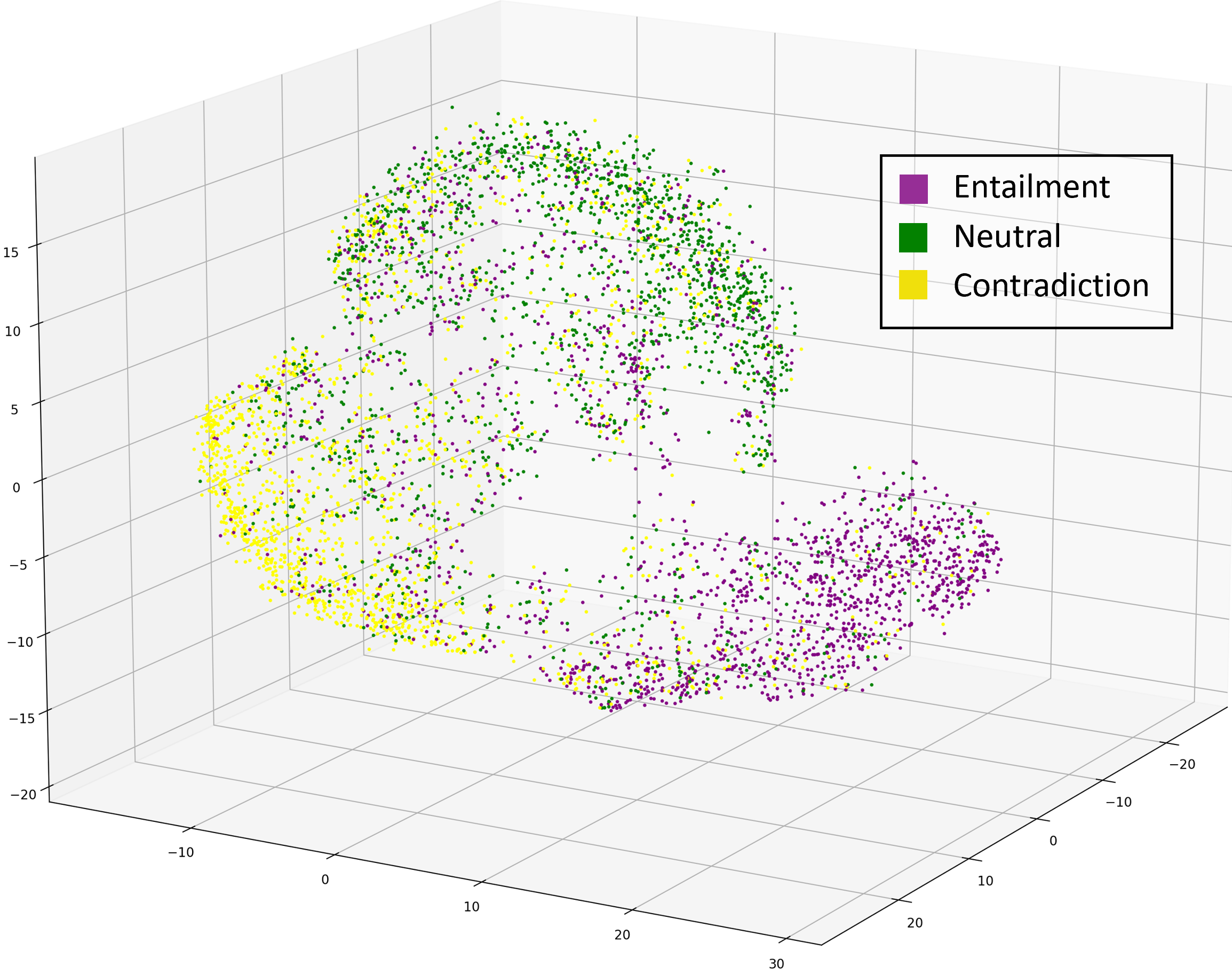}
  \subcaption{Class-wise Labels (Source: {\small EN})}
\endminipage\hfill\minipage{0.31\textwidth} \includegraphics[width=\linewidth]{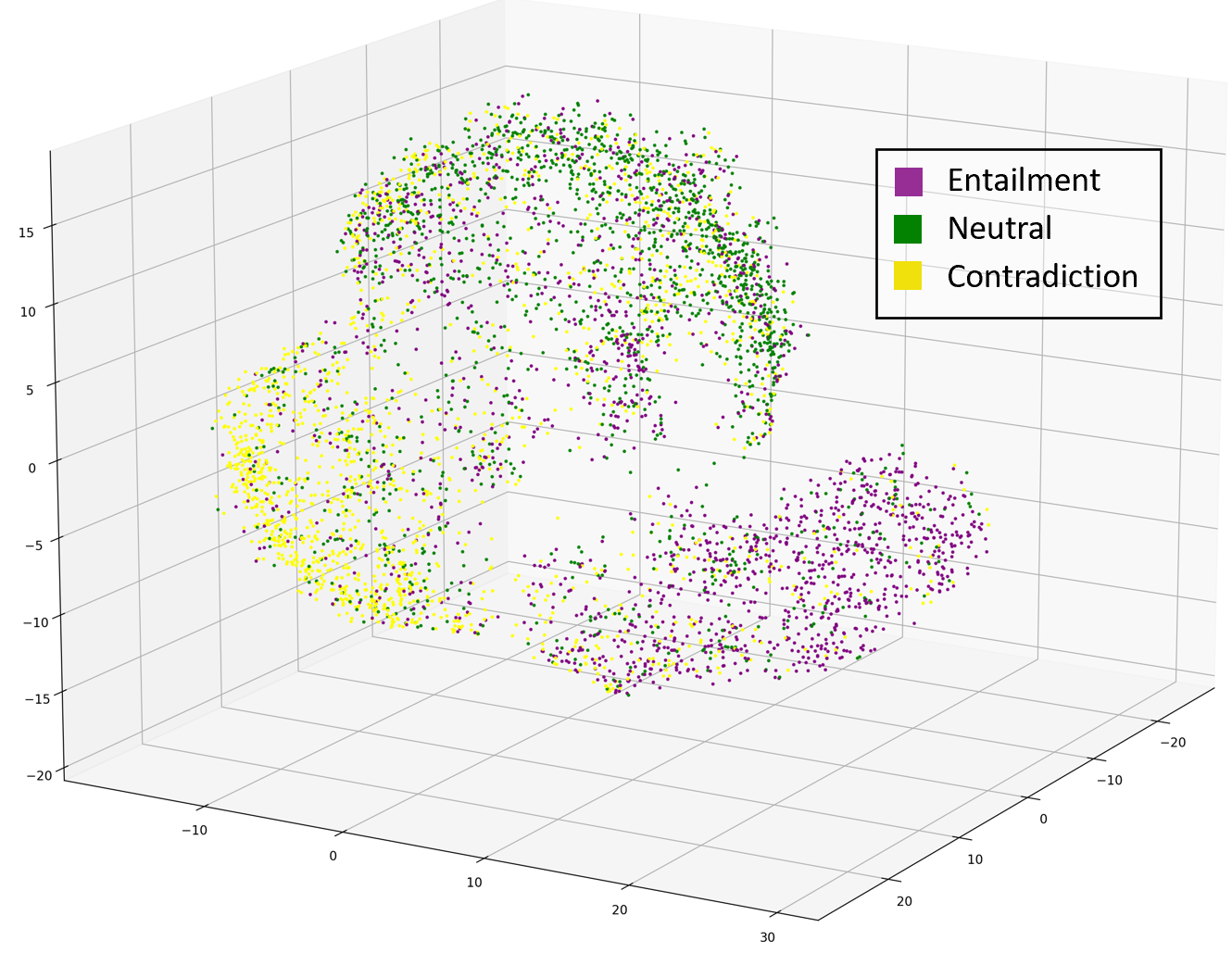}
  \subcaption{Class-wise Labels (Target: {\small RU})}
\endminipage
\end{subfigure}
\vspace{-0.7em}
\caption{3D t-SNE visualization of the features from the last layer of   fine-tuned mBERT on XNLI ($S$=1\%).}
\label{fig:ditto_3dtsne}
% \vspace{-1em}
\end{figure*}
% \vspace{-1em}
\hspace{-1em}\textbf{Sharpness-Aware Minimization (SAM)}:
SAM seeks to find the parameter $w$ such that even its neighborhood has seemingly similar low training loss $\mathcal{L}_{\text{train}}$ with minimal variation by optimizing the following objective:
% which lie in a neighborhood having uniformly low training loss $\mathcal{L}_{\text{train}}$ by optimizing the following objective:
\vspace{-0.5em}
\begin{equation}
\begin{aligned}
\min_{w} \quad & \max_{{||\epsilon||}_2 \leq \rho} & \mathcal{L}_{\text{train}} (w+\epsilon)   \\
\end{aligned}
\end{equation}
where $\rho$ is the size of the neighborhood. Since, the exact solution of the inner maximization is hard to obtain, the authors of SAM propose a simple first order approximation:
\vspace{-0.5em}
\begin{equation}
\begin{aligned}
\hat{\epsilon(w)} & \approx \\ \argmax_{{||\epsilon||}_2 \leq \rho} & \, \mathcal{L}_{train}(w) + \epsilon^T \nabla_w \mathcal{L}_{\text{train}}(w) \\ 
& = \rho\nabla_w\mathcal{L}_{\text{train}}(w)/ ||\nabla_w\mathcal{L}_{\text{train}}(w)||_2  \\
\end{aligned}
\end{equation}
After computing $\hat{\epsilon}$, the parameter $w$ is updated based on the the sharpness-aware gradient $\nabla_w\mathcal{L}_{\text{train}}(w)|_{w+\hat{\epsilon(w)}}$. 

\hspace{-1em}\textbf{Domain-Adversarial Neural Networks (DANN)}:
DANN \cite{ganin2016domain} has been
successful applied for many unsupervised domain adaptation tasks for minimizing the domain shift \cite{du2020adversarial, long2018conditional}. DANN needs a labeled source domain dataset $\mathbb{L}_s = \{(x^s_i, y^s_i)\}^{n}_{i=1}$ with $n$ examples and an unlabeled target domain dataset $\mathbb{U}_t = {\{x^t_i\}}^{m}_{i=1}$ with $m$ examples. It consists of three modules: Encoder $\mathcal{E}$, Task-Specific Classifier $\mathcal{C}$, and Domain Discriminator $\mathcal{D}$. In a nutshell, DANN requires solving a two-player  game where the first player is the Domain Discriminator $\mathcal{D}$, is trained to distinguish the target domain from the source domain, and the second player is the encoder $\mathcal{E}$, which is trained simultaneously to confuse the Discriminator $\mathcal{D}$ such that the encoder learns to generate domain invariant features. 
% Optimization of DANN requires two losses terms: (a) task-specific classification loss $\mathcal{L_C}$ and (b) domain classification loss $\mathcal{L_D}$.
We minimize the task-specific classification loss $\mathcal{L_C}$ using the source domain labeled dataset for optimizing the classifier $\mathcal{C}$ and encoder $\mathcal{E}$.
\vspace{-0.5em}
\begin{equation}
\begin{aligned}
\mathcal{L_C} (\mathcal{E}, \mathcal{C}) = \mathbb{E}_{{(\textbf{x}^s_i, \textbf{y}^s_i) \sim \mathbb{L}_s}}\mathcal{L}(\mathcal{C}(\mathcal{E}(\textbf{x}^s_i)), \textbf{y}^s_i)   \\
\end{aligned}
\end{equation}
\vspace{-1em}

$\mathcal{D}$ is trained to predict the domains by minimizing domain classification loss:
\vspace{-0.5em}
\begin{equation}
\begin{aligned}
\mathcal{L_D}(\mathcal{E}, \mathcal{D}) = -\mathbb{E}_{{\textbf{x}^s_i \sim \mathbb{L}_s}}\log[\mathcal{D}(\mathcal{E}(\textbf{x}^s_i))] \\ -\mathbb{E}_{{\textbf{x}^t_j \sim \mathbb{U}_t}}\log[1 - \mathcal{D}(\mathcal{E}(\textbf{x}^t_j))]   \\
\end{aligned}
\end{equation}
$\mathcal{L}_D$ is maximized for $\mathcal{E}$ so that $\mathcal{D}$ is not able to distinguish between the domains. The minimax optimization of DANN is defined as:
\vspace{-0.5em}
\begin{equation}
\begin{aligned}
\min_{\mathcal{E}, C}  \quad & \mathcal{L}_C (\mathcal{E}, C) -
\lambda \mathcal{L_D} (\mathcal{E}, \mathcal{D})\\
\min_{\mathcal{D}}  \quad & \mathcal{L_D} (\mathcal{E}, \mathcal{D})
\end{aligned}
\end{equation}
where $\lambda$ is a hyper-parameter to control trade-off between classification and domain adversarial loss.

\section{Limitations of Zero-shot Learning}
\begin{table}[ht]
\centering
\small
\setlength\tabcolsep{4pt}%
\begin{tabular}{cc|ccc|ccc}
\toprule
\multicolumn{1}{c}{} & \multicolumn{1}{c}{} & \multicolumn{3}{c}{mBERT}  & \multicolumn{3}{|c}{XLM-R} \\
    \cmidrule(lr){3-5}\cmidrule(lr){6-8}
 \multirow{-1}{*}{\textbf{Dataset}} & \multirow{-1}{*}{\textbf{|$\mathbb{T}$|}} & 1\%  & 10\%&  100\% &1\%  & 10\%&  100\% \\ 
\midrule
XNLI & 14 & 10.3 & 12.3 & 15.5 & 8.3 & 10.3 & 11.3 \\
MARC & 5 & 14.8 & 18.1 & 20.3 & 4.5 & 8.8 & 9.9  \\
AmNLI & 10 & {\color{red} 24.8} & {\color{red} 32.9} & {\color{red} {41.2}} & {\color{red} 29.9} & {\color{red} 39.2} & {\color{red} {45.1}} \\
% PAWS-X & 6 & 9.9 & 12.9 & 12.0 & 4.2 & 10.9 & 11.6 \\
\bottomrule
\end{tabular}
\vspace{-0.5em}
\caption {We have reported the mean of difference $\triangle$ between the zero-shot performance of all the target languages and source language for varying amount of the source language ({\small EN}) data used while fine-tuning. |$\mathbb{T}$| is the number target languages available in the dataset.} 
\label{tbl:zeroshot_delta}
\end{table}

\textbf{Inconsistent Cross-Lingual Transfer}:
We have reported the average difference ($\delta$) in the zero-shot performance between the target and the source language in Table \ref{tbl:zeroshot_delta}. 
We experiment with mBERT and XLM-R on XNLI \cite{conneau2018xnli}, AmNLI \cite{ebrahimi-etal-2022-americasnli} and MARC \cite{keung-etal-2020-multilingual} datasets to measure the average $\delta$ between zero-shot performance of the target and source language. Table \ref{tbl:zeroshot_delta} shows that XNLI and AmNLI having relatively higher $\delta$ due to diverse number of languages. We also notice that mBERT has a higher $\delta$ than XLM-R across all tasks except AmNLI, showing the importance of amount of pre-training size.
% We observe significant deltas across datasets, with XNLI \cite{conneau2018xnli} and AmNLI \cite{ebrahimi-etal-2022-americasnli} having relatively higher delta due to diverse number of languages. We also notice that mBERT has a higher delta than XLM-R across all tasks, showing the importance of amount of pre-training size. 

\hspace{-1em}\textbf{Feature Incongruity between Languages}: We hypothesize that the inconsistent zero-shot performance is due to the mismatch in the feature representation space of the fine-tuned MMT on the source language. To verify that we visualize the target and source language feature representations learned using standard zero-shot training method  using 3D t-SNE \cite{van2008visualizing} in Figure \ref{fig:ditto_3dtsne}.
In Figure \subref{fig:zeroshot_language}, there is clear distinction between the source (En) and target language (Ru) features. While in Figure \ref{fig:zeroshot_source} and \ref{fig:zeroshot_target}, the feature space for the entailment class is overlapping with the source language, but fairly distinct for the other two classes, this could be potential cause for inferior cross-lingual transfer.
% This can lead towards more false positive entailment for the classification model, affecting the cross-lingual transfer.

We measure centered kernel alignment (CKA) \cite{kornblith2019similarity} between the source and the target language feature representations to quantify the incongruity. In Figure \ref{fig:baseline_cka}. We have plotted the CKA similarity with the zero-shot performance across all the languages. The plot suggests that there is a strong correlation between CKA and zero-shot performance, with Pearson and Spearman correlation coefficients as 0.98 and 0.96, respectively establishing our hypothesis.  

\section{Unveiling \DITTO}
Typological similarity and incongruency between feature representations lead us to envision different languages as domains. As discussed in the previous section, DANN is useful in minimizing the domain shift across domains using only unlabeled data in the target domain. We propose to perform  adversarial adaptation of the target language features for transforming the same towards the source language feature distribution. 

We have a set of target languages $\mathbb{T}$ with each target language $t$ having dataset  $\mathbb{U}_t = \{x^t_i\}^{\mathcal{T}}_{i=1}$ with $\mathcal{T}$ unlabeled examples and an unlabeled set $\mathbb{U}_s = \{x^s_i\}^{\mathcal{S}}_{i=1}$ with $\mathcal{S}$ examples in the source language. In DANN, there is one target domain, whereas in our case we have a set of target languages $\mathbb{T}$ and we hypothesize and empirically show that performing adaptation for each language separately may cause degradation in other target languages, as seen in Table \ref{tbl:ditto_lang_sampling}. Hence, we propose \DiTTO where we jointly perform adaptation across all target languages. 

\DiTTO consists of an MMT $\mathcal{M}$ for encoding the features, a task-specific classifier $\mathcal{C}$ and Language Discriminators $\mathcal{D}^L = \{\mathcal{D}^L_t\}^{|\mathbb{T}|}_{i=1}$. We train these modules using two losses: task-specific classification loss $\mathcal{L_C}$, defined in the Equation (\ref{eqn:standard_zeroshot}) and language discrimination loss $\mathcal{L}_\mathtt{L}$ for distinguishing the target  and source language. 

As we have $|\mathbb{T}|$ discriminators, we randomly sample a target language $t$ from a prior distribution $p(\mathbb{T})$ at each training step and train the discriminator $ \{\mathcal{D}^L_t\}$ to accurately distinguish target $t$ and source language using the following loss:
\vspace{-0.5em}
\begin{equation}
\begin{aligned}
\mathcal{L}_L (\mathcal{M}, \textbf{D}^L_t) = -\mathbb{E}_{{\textbf{x}^s_i \sim \mathbb{U}_s}}\log[b^L_t(\mathcal{M}(\textbf{x}^s_i))] \\ -\mathbb{E}_{{\textbf{x}^t_j \sim \mathbb{U}_t}}\log[1 - \mathcal{D}^L_t(\mathcal{M}(\textbf{x}^t_j))]   \\
\end{aligned}
\end{equation}
We maximize the above loss $\mathcal{L}_L (\mathcal{M}, \mathcal{D}^L_t)$ for confusing the language discriminator $\mathcal{D}^L_t$ to transform the target features towards the source language.

In our initial experiments (reported in Table \ref{tbl:ditto_ablation}), we observed some instability due to adversarial adaptation \cite{mao2017least, xing2021algorithmic}. We propose optimizing the task-specific loss $\mathcal{L}_C$ using SAM so that it may generalize to the target languages, improving the stability during adversarial adaptation. We directly fine-tune the MMT $\mathcal{M}$ on the source language labeled dataset $\mathcal{D}^l_s$ by minimizing Equation (\ref{eqn:standard_zeroshot}) using SAM. Following DANN and SAM, the final optimization objective of \DiTTO  can be defined as:
\vspace{-0.5em}
% \begin{mdframed}[backgroundcolor=purple!10]
% \begin{multline}
% \min_{\mathcal{M}, C} \quad \max_{{||\epsilon||}_2 \leq \rho}  \mathcal{L}_{C}(\hat{\mathcal{M}}, \hat{C})\\ 
% -\lambda \mathbb{E}_{t \sim p(\mathbb{T})} \mathcal{L}_L(\mathcal{M}, \mathcal{D}^L_t)
% \end{multline}
% \begin{equation}
%     \min_{\mathcal{D}_L}  \quad \mathbb{E}_{t \sim p(\mathbb{T})} \mathcal{L}_L(\mathcal{M}, \mathcal{D}^L_t)
% \end{equation}
% \end{mdframed}
% \begin{mdframed}[backgroundcolor=purple!10]
\begin{gather} 
\min_{\mathcal{M}, C} \max_{{||\epsilon||}_2 \leq \rho}  \mathcal{L}_{C}(\hat{\mathcal{M}}, \hat{C})  -\lambda \mathbb{E}_{t \sim p(\mathbb{T})} \mathcal{L}_L(\mathcal{M}, \mathcal{D}^L_t) \label{eqn:mmt_opt}
\\
\vspace{-0.5em}
\min_{\mathcal{D}_L}  \quad \mathbb{E}_{t \sim p(\mathbb{T})} \mathcal{L}_L(\mathcal{M}, \mathcal{D}^L_t)  \label{eqn:dis_opt}
\end{gather} 
% \end{mdframed}
% \vspace{-1em}
where $\hat{\mathcal{M}}$, $\hat{C}$ are the updated parameters using $\epsilon$.

% \subsection{Sharpness-Aware Optimization Improves Cross-Lingual Transfer}

% \subsection{Language Adversarial Training}
% \cite{lauscher2020zero, kumar-etal-2022-diversity} observes that the zero-shot performance varies across languages  depending upon two major factors i) typological similarity between the source and target language and ii) size of the target language during pre-training. Assuming the target language has sufficient data during pre-training, zero-shot transfer will be only influenced by the typological similarity between the source and target language.

% CKA \cite{kornblith2019similarity}
% Typological similarity can be seen as the distributional shift from the source and target language. 

\section{Experimental Setup}
\subsection{Datasets}
We evaluate our method on three benchmark datasets consisting of languages from various language families, to ensure better cross-lingual transfer evaluation. \textbf{XNLI} dataset \cite{conneau2018xnli} consists of translated dataset in 14 languages from English. The task requires any model to predict whether the premise entails, contradicts, or neutral to the given hypothesis. AmericasNLI (\textbf{AmNLI}) dataset \cite{ebrahimi-etal-2022-americasnli} is an extension of XNLI to 10 indigenous languages of the Americas, which are even unseen during pre-training of XLM-R and mBERT. Multilingual Amazon Review Corpus (\textbf{MARC}) dataset \cite{keung-etal-2020-multilingual} is a large-scale dataset consisting of Amazon reviews for text classification in 6 languages. We use the review text and title to predict its star rating. 
% \textbf{PAWS-X} dataset \cite{yang2019paws} consists of human-translated paraphrase identification pairs from English language to six typologically distinct languages. The task requires any classifier to predict whether two sentences are paraphrases of each other.

% \vspace{0.5em}
% \hspace{-1em}\textit{Cross-lingual Natural Language Inference} (XNLI). The XNLI dataset \cite{conneau2018xnli} consists of translated dataset in 14 languages of English hypothesis-premise pairs.

% \vspace{0.5em}
% \hspace{-1em}\textit{AmericasNLI} (AmNLI). The AmNLI dataset \cite{ebrahimi-etal-2022-americasnli} is an extension of XNLI to 10 indigenous language of the Americas which are even unseen during pretraining of XLM-R and mBERT.

% \vspace{0.5em}
% \hspace{-1em}\textit{Multilingual Amazon 
% Review Corpus} (MARC). The MARC dataset \cite{keung-etal-2020-multilingual} is large-scale dataset consisting of Amazon reviews for text classification in 6 languages. We use the review text and title for predicting it's star rating.

% \vspace{0.5em}
% \hspace{-1em}\textit{Cross-lingual Paraphrase Adversaries from Word
% Scrambling} (PAWS-X). The PAWS-X dataset \cite{yang2019paws} consists of human translated paraphrase identification pairs from English language to six typologically distinct languages.

\subsection{Baselines and \textit{DiTTO} Variants}
In the \textbf{Baseline} experiments, we fine-tune MMTs on labeled data of the source language using Equation (\ref{eqn:standard_zeroshot}). In the vanilla \textbf{\DiTTO} setup, we use all the target languages available in the dataset. In the vanilla setup, we want to assign a higher probability to those target languages with a lower zero-shot performance from the Baseline method. We defined the prior distribution $p(\mathbb{T})$ of target languages as follows:
\vspace{-0.5em}
\begin{gather}
    \Delta_t = \max(\mathcal{Z}(s) - \mathcal{Z}(t), 0) \\
    p(t) = \delta_t + \sigma_{\Delta_t}
\end{gather}
where, $\mathcal{Z}$ is the zero-shot performance from the Baseline method, $\Delta_t$ is the non-negative delta between the source and target language, and $\sigma_{\delta}$ is the standard deviation of the $\Delta_t$ across all the target languages.
 
\hspace{-1em}\textbf{\textit{DiTTO (UNF)}} is a variant of vanilla {\DiTTO} in which we set the prior distribution $p(\mathbb{T})$ to be uniform across all the target languages. \textbf{\textit{DiTTO} (t)} is a single target language variant of {\DiTTO} where only one target language $t$ is used during training. \textbf{\textit{DiTTO}-LA} does not perform adaptation of the target languages, however optimization is done using SAM on the source language labeled data. \textbf{\textit{DiTTO}-SAM} performs language adaptation without SAM.
% \textbf{\textit{DiTTO} + MCC} is another variant where MCC is used along with adversarial training for improving the adaptation of target languages.

\subsection{Training Details}
We conduct all of our experiments using mBERT (\textit{bert-base-multilingual-cased}) and XLM-R (\textit{xlm-roberta-base}). We use a batch size of 32 and a maximum sequence length of 128 across all the datasets. We fine-tune for $\{15, 20, 25\}$, $\{3, 5, 7\}$, $\{2, 3, 5\}$ epochs while using 1\%, 10\% and 100\% of the source language data respectively. We use the AdamW \cite{loshchilov2018decoupled} optimizer with linear scheduler and learning rate as 1e-5 for the encoder and classifier and 5e-5 for the discriminator. We set the $\lambda$ hyper-parameter as 1 for all the experiments. We run experiments for each hyper-parameter and report the best average accuracy on three random seeds. 

\section{Results}
In this section, we describe the results of several experiments to analyze the \DiTTO method and compare its performance with the Baseline in the zero-shot setting. In order to justify the robustness of our method, we conduct experiments with the varying amount of source language data.
% We perform these experiments to investigate our method's performance when there are few source language labeled data with the very low cross-lingual transfer. 
In our experiments, {\small EN} is the default source language and we categorize target languages as follows: 
1. \textit{Distant}: languages that are typologically dissimilar to the source language
2. \textit{Low-resource}: languages that have scarcity of data for pre-training
3. \textit{Unseen}: languages that were not included in the pre-training data of MMT.
Furthermore, we compare the techniques in the few-shot setting with few labeled examples in target languages. Then, we perform a thorough ablation study and analyze various variants of \DiTTO. Finally, we show evidence in the form of congruity between the source and target language feature representations and t-SNE visualization in support of our hypothesis.

\subsection{Zero-shot Transfer Results}
%We compare \DiTTO over Baseline in the zero-shot setting across various datasets, target languages and amount of source language data.
\textbf{Performance across datasets}: In Table \ref{tbl:relative_gain_ditto}, we have reported the relative gains  from \DiTTO for zero-shot setting averaged across all the languages over the baseline method using 1\%, 10\% and 100\% of the source language data. We observe the gains are positive (upto 23.05\%) across all the training configurations. The gains are much higher for mBERT than XLM-R due to lower cross-lingual transfer in mBERT except the AmNLI dataset. The relative gains start to decrease with the increased amount of the source language data $S$ on all the datasets except AmNLI, where the gains remains consistent for higher values of $S$ (10\% and 100\%).
% We also measured the efficacy of \DiTTO using another source language (ZH) to ensure that \DiTTO works independent to the source language (cf. Table \ref{tbl:relative_gain_ditto_zh}).

\begin{table}[ht]
\centering
\small
\setlength\tabcolsep{4pt}%
\begin{tabular}{c|ccc|ccc}
\toprule
\multicolumn{1}{c}{} & \multicolumn{3}{c}{mBERT}  & \multicolumn{3}{|c}{XLM-R} \\
    \cmidrule(lr){2-4}\cmidrule(lr){5-7}
 \multirow{-1}{*}{\textbf{Dataset}} & 1\%  & 10\%&  100\% &1\%  & 10\%&  100\% \\ 
\midrule
XNLI & \textbf{23.05} & 6.58 & 2.10 & 13.57 & 4.10 & 2.71 \\
AmNLI & 11.61 & \textbf{19.72} & 15.10 & 17.95 & \textbf{19.87} & \textbf{19.09}  \\
MARC & 12.28 & 15.40 & \textbf{19.03} & 5.61 & 3.04 & 2.41 \\
% PAWS-X & 14.10 & \textbf{16.87} & 2.19 & 3.49 & 0.87 & 0.44 \\
\bottomrule
\end{tabular}
\vspace{-0.5em}
\caption {Relative gains (in \%) of \DiTTO over Baseline.} 
\label{tbl:relative_gain_ditto}grey
\end{table}

\begin{figure}[ht]
\centering
\includegraphics[scale=0.17]{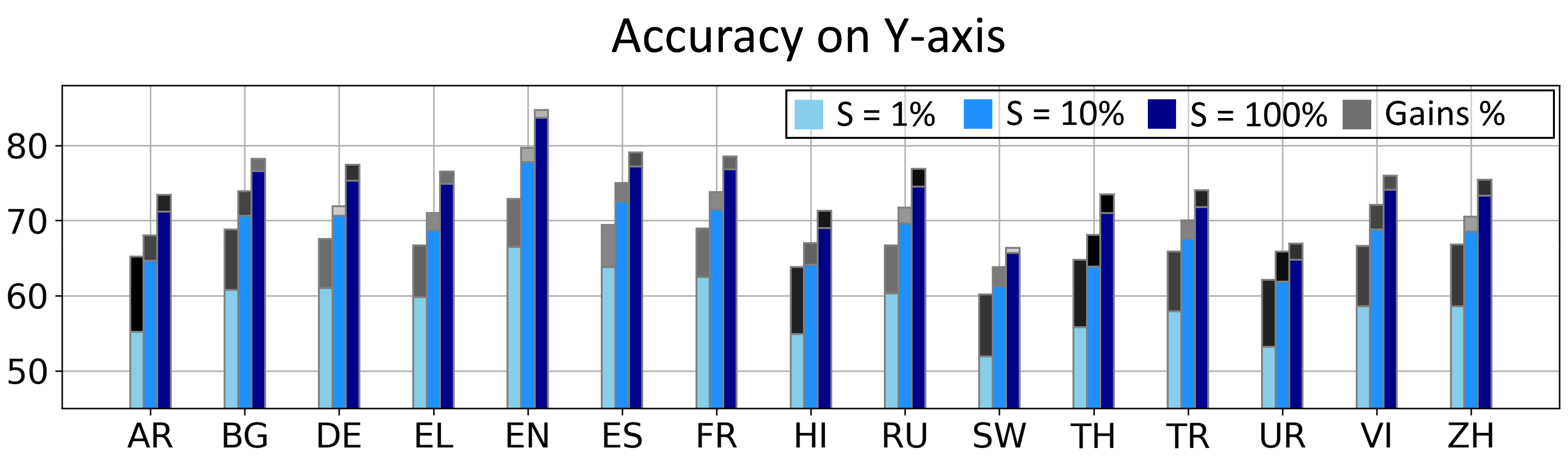}
\vspace{-0.5em}
  \caption{Absolute gains (darker shades of grey denotes higher gains) from \DiTTO for XLM-R on XNLI dataset. Magnified view available in Figure \ref{fig:language_xnli} in Appendix.}
    \label{fig:language_xnli_xlmr}
    \vspace{-0.5em}
\end{figure}

\hspace{-1em}\textbf{Performance across Seen Target Languages}: We have reported the absolute gains of \DiTTO in Figure \ref{fig:language_xnli_xlmr} on XNLI using XLM-R 
%and provided similar analysis on all the datasets (XNLI and MARC) using both the MMTs in Figures \ref{fig:language_xnli} and \ref{fig:language_marc} of the Appendix.
We observe positive gains from \DiTTO for all the target languages, with much larger gains especially on the low-resource and distant languages compared to the Baseline model. Similar to the earlier observation in Table \ref{tbl:relative_gain_ditto}, the gains starts to decrease across target languages as we increase the amount of the source language data.

\begin{figure}[!]
\centering
\includegraphics[scale=0.18]{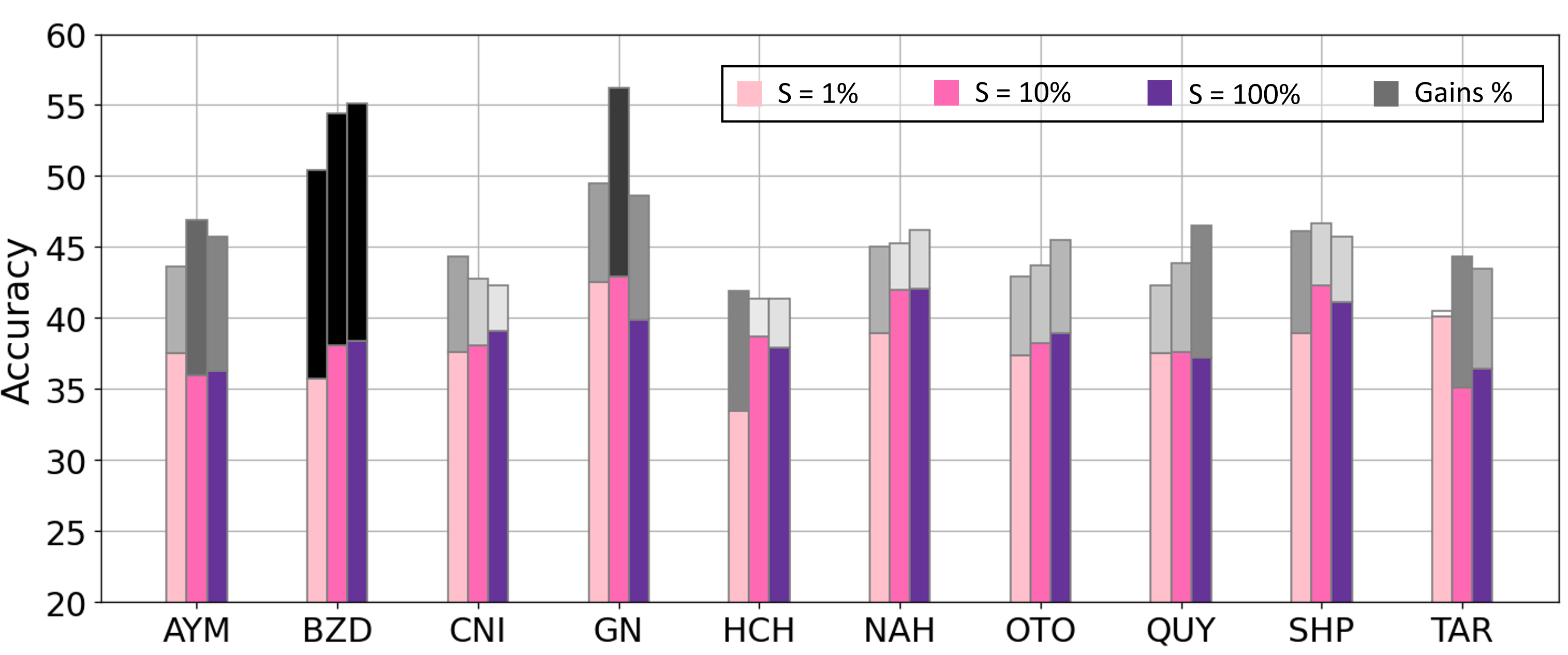}
\vspace{-0.5em}
  \caption{Absolute gains (darker shades of grey denotes higher gains) from \DiTTO for XLM-R on AmNLI.}
    \label{fig:language_amli_xlmr}
    \vspace{-0.5em}
\end{figure}

\hspace{-1em}\textbf{Performance across Unseen Target Languages}: To measure the impact of \DiTTO on unseen languages, we report the absolute gains from \DiTTO on XLM-R on the AmNLI dataset in Figure \ref{fig:language_amli_xlmr}. We have provided a similar analysis for mBERT in Figure \ref{fig:language_amli} of the Appendix. The gains from \DiTTO are consistent across all unseen languages. We observe that the gains are higher for languages with better Baseline performances, which is in contrast to trends on seen languages. For unseen languages, we do not observe the trend of diminishing gains with an increase in the source language data. If we compare the gains on AmNLI with the XNLI dataset, we notice \DiTTO providing on average 1.7 times higher gains across all the configurations.

% \subsection{Comparing DiTTO and Baseline performances in Few-shot setting}
\subsection{Few-shot Transfer Results}
\begin{figure}[!h]
\centering
\includegraphics[scale=0.195]{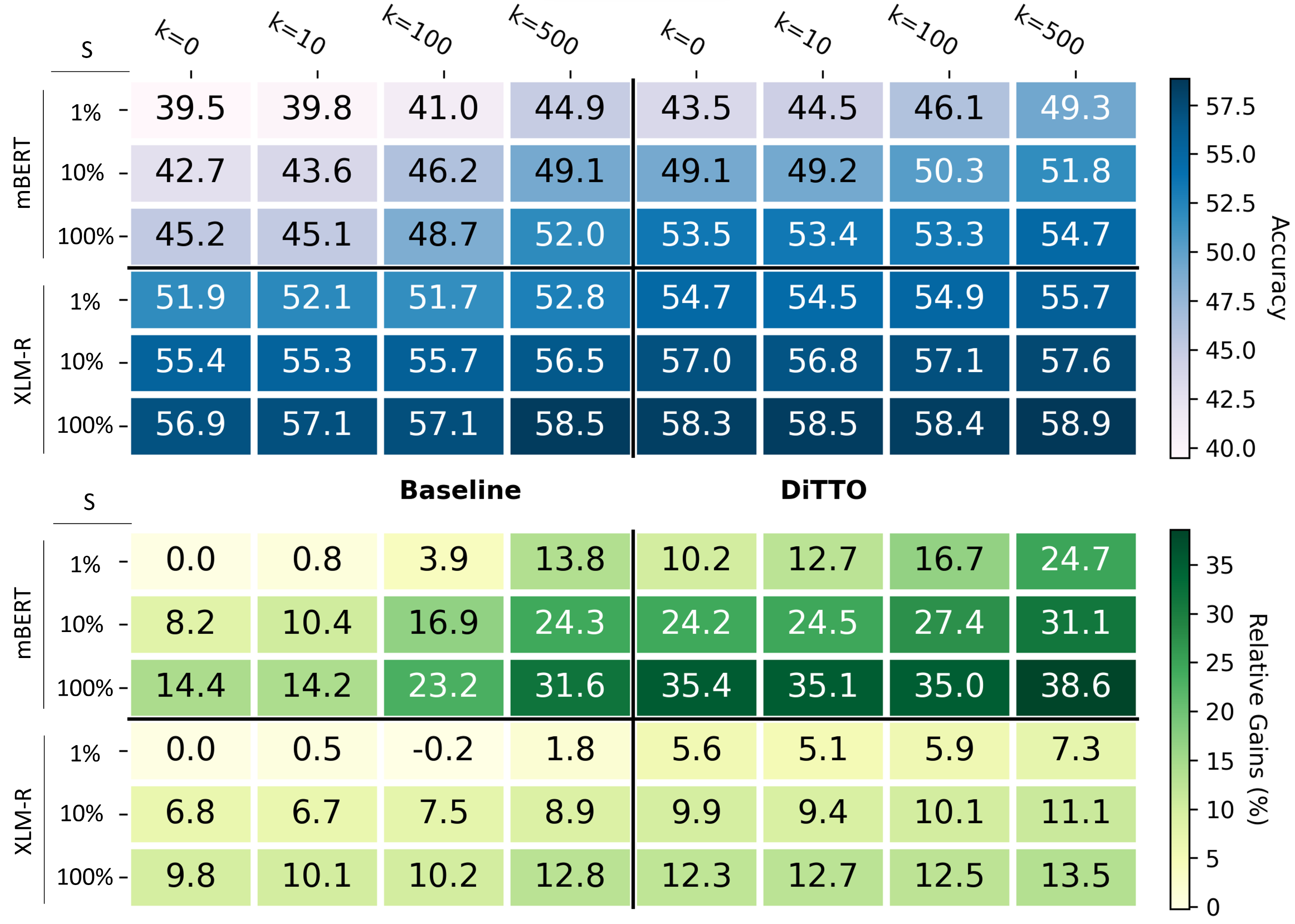}
\caption{Accuracy/relative gains on MARC dataset. 
% Rows and columns denoting the amount of source and target language labeled instances, respectively.
% The bottom heat-map reports the relative gains from zero-shot accuracy using Baseline with 1\% of the source language data.
 }
\label{fig:heatmap_fewshot_marc}
\vspace{-1em}
\end{figure}

It can be argued that the gains from \DiTTO in zero-shot setting can be achieved using few-shot cross-lingual transfer. Therefore, we conduct experiments in the few-shot setting by adding $k$ labeled instances in each of the target languages to measure capabilities of \DiTTO when some labeled data is available along with unlabeled data. In Figure \ref{fig:heatmap_fewshot_marc}, we have reported the accuracy and relative gains\footnote{The relative gain is calculated with respect to the accuracy obtained by the Baseline method on $S=1\%$ and $k=0$.} using Baseline and \DiTTO on MARC dataset. We have provided a similar analysis for AmNLI dataset in Figure \ref{fig:heatmap_fewshot_amnli} in the Appendix. 
% These heat maps show that when the performance is low using Baseline, the gains from \DiTTO are higher.

The heat maps in Figure \ref{fig:heatmap_fewshot_marc} show that while XLM-R has better accuracy than mBERT for both Baseline and \DiTTO setup, but the gains (both absolute and relative) on mBERT for both methods are higher compared to XLM-R. We also notice that by increasing either the source or the target language data, performance for both Baseline and DiTTO increased and hence, we will compare the gains of \DiTTO with Baseline in these two dimensions.

\hspace{-1em}\textbf{Impact of Target Language Labeled Data}: We observe that when we fix the amount of source language data and increase the value of $k$, the gains from \DiTTO are higher than Baseline. Also, the gains from \DiTTO on $k$=0 is comparable with the gains of baseline on $k$=500. In AmNLI, the gains from \DiTTO for lower values of $k$ are quite high compared to baseline, while for higher values of $k$ Baseline performance for XLM-R is comparable to DiTTO.

\hspace{-1em}\textbf{Impact of Source Language Labeled Data}: We also noticed that by fixing the value of $k$ and increasing the size of source language data $S$, there is an increase in gains for both methods on MARC. However, the increase in gains from \DiTTO is much higher than Baseline. At the same time, on the AmNLI dataset consisting of unseen target languages, the gains is much smaller with the increase in $S$ (cf. Appendix).

\hspace{-1em}\textbf{Chinese as Source Language}: To measure the effectiveness of \DiTTO across different source languages, we conduct zero-shot experiments considering Chinese ({\small ZH}) as the source language on the MARC dataset. We have reported the average accuracy across all the languages in Table \ref{tbl:relative_gain_ditto_zh}. \DiTTO provides consistent gains over the Baseline method across all the training configurations, comparatively higher gains than {\small EN} as the source language. 

\begin{table}[ht]
\centering
\setlength\tabcolsep{4pt}
\small
\begin{tabular}{c|ccc|ccc}
\toprule
\multicolumn{1}{c}{} & \multicolumn{3}{c}{mBERT}  & \multicolumn{3}{|c}{XLM-R} \\
    \cmidrule(lr){2-4}\cmidrule(lr){5-7}
 \multirow{-1}{*}{\textbf{Dataset}} & 1\%  & 10\%&  100\% & 1\%  & 10\%&  100\% \\ 
\midrule
Baseline & 32.88 & 39.48 & 42.68 & 45.83 & 50.86 & 51.38 \\
\DITTO & \textbf{39.09} & \textbf{46.90} & \textbf{50.82} & \textbf{51.84} & \textbf{53.34} & \textbf{55.27} \\
\midrule
RG(\%) & \textbf{19.45} & \textbf{20.80} & \textbf{20.67} & \textbf{13.31} & 5.34 & 8.12 \\
\bottomrule
\end{tabular}
\vspace{-0.5em}
\caption {We have reported the zero-shot accuracy averaged across all languages with \textbf{{\small ZH}} as the source language data on MARC dataset. RG denotes the relative gains averaged across all the languages from using \DITTO over Baseline.} 
\label{tbl:relative_gain_ditto_zh}
\end{table}

\hspace{-1em}\textbf{Performance and Cost Trade-off}:
\DiTTO is seven times more cost-effective in terms of both source and target language data.
% \footnote{Across various values of source language data ($S$) and few-shot target language data ($k$)}. 
We validate this by plotting the accuracy from both methods against the cost incurred while collecting the labeled data for fine-tuning. For detailed analysis refer to the section \ref{section:cost} in the Appendix.

\subsection{Ablations and Variants Analysis}
\label{section:ablation}
% We deep dive all the components of our model to uncover what actually makes it work, and why, by performing three studies:
% We analyze our model to see where the improvement is coming from, which parts are working, and why with the help of three studies.

\textbf{Ablation Study}:
Here we scrutinize the contributions from adaptation of target languages and optimization with SAM. We report the zero-shot relative gains in Table \ref{tbl:ditto_ablation} by ablating each of these components. We observe that removing any component reduces the performance for most of the training configurations, indicating that both target language adaptation and optimization have a contribution in achieving better results. We also observe that removing SAM (\textit{DiTTO} - SAM) leads to unstable performances on XNLI and AmNLI datasets with negative relative gains on AmNLI ($S$=1\%) for both MMTs, and on XNLI ($S$=10\%) for mBERT, showing instability caused in adversarial training \cite{mao2017least,xing2021algorithmic}. Removing target language adaptation (\textit{DiTTO}-LA) reduces the relative gains by a significant margin, showing the importance of adaptation of target language features. It performs similar to \DiTTO on XNLI ($S$=1\%) dataset using mBERT, demonstrating just optimization using SAM can also improve cross-lingual transfer. The performance of \textit{DiTTO} - SAM, it is often higher compared to \textit{DiTTO} - LA, which indicates that Language Adaptation is a much more crucial for improving cross-lingual transfer.

\begin{table}[!h]
\centering
\small
\setlength\tabcolsep{2pt}%
\begin{tabular}{c|c|ccc|ccc}
\toprule
\multicolumn{1}{c}{} &  &\multicolumn{3}{c}{mBERT}  & \multicolumn{3}{|c}{XLM-R} \\
    \cmidrule(lr){3-5}\cmidrule(lr){6-8}
 \multicolumn{1}{c}{} &\multirow{-1}{*}{\textbf{Method}} & 1\%  & 10\%&  100\% & 1\%  & 10\%&  100\% \\ 
\midrule
& \DITTO & \textbf{23.05} & \textbf{6.58} & \textbf{2.10} & \textbf{13.57} & \textbf{4.10} & \textbf{2.71} \\
& \textit{DiTTO} - SAM & 8.84 & {\color{red}-0.36} & 1.80 & 6.04 & 1.81 & 2.02 \\
\multirow{-3}{*}{\rotatebox[origin=c]{90}{XNLI}} & \textit{DiTTO} - LA & {22.25} & 3.89 & 2.02 & 7.43 & 2.81 & 1.74 \\
\midrule
& \DITTO & \textbf{12.28} & \textbf{15.40} & \textbf{19.03} & \textbf{5.61} & \textbf{3.05} & \textbf{2.41} \\
& \textit{DiTTO} - SAM & 8.64 & 9.90 & 14.98 & 2.89 & {\color{red}-0.13} & 2.27 \\
\multirow{-3}{*}{\rotatebox[origin=c]{90}{MARC}} & \textit{DiTTO} - LA & 5.5 & 1.54 & 2.20 & 4.02 & 0.35 & {\color{red}-0.54} \\
\midrule
& \DITTO & \textbf{11.61} & \textbf{19.72} & \textbf{15.10} & \textbf{17.95} & \textbf{19.87} & \textbf{19.09}
\\
& \textit{DiTTO} - SAM & {\color{red}-3.85} & 14.35 & 14.52 & {\color{red}-11.88} & 14.81 & 15.89 \\
\multirow{-3}{*}{\rotatebox[origin=c]{90}{AmNLI}} & \textit{DiTTO} - LA & 7.21 & 5.17 & {\color{red}-1.00} & 7.57 & 7.58 & 9.33  \\
\bottomrule
\end{tabular}
\vspace{-0.5em}
\caption {Ablation Study: Zero-shot relative gains (in \%) averaged across all the languages over Baseline.} 
\label{tbl:ditto_ablation}
\end{table}

\hspace{-1em}\textbf{Single vs Multiple Target Language Adaptation}: In the base setup of \DiTTO, we propose to perform an adaptation of all the target languages available in the dataset. We conduct zero-shot experiments with a single target language variant \textit{DiTTO} (t) to validate our assumption. In Table \ref{tbl:ditto_lang_sampling}, we observe that the single language variant provides similar gains as the vanilla \DiTTO in the selected language $t$. However, often there is very little/no improvement observed in languages other than $t$. \textit{DiTTO} (JA) and \textit{DiTTO} (ZH) under-perform than Baseline for most of the languages.

\begin{table}[ht]
\centering
\small 
\setlength\tabcolsep{3pt}%
\begin{tabular}{c|cccccc|c} 
\toprule
Method &  {\small EN}  & {\small DE}  & {\small ES}      & {\small FR}  & {\small JA}  & {\small ZH} &  AVG\\
\midrule 
Baseline & 54.3	& 42.3	& 42.3	& 43.8	& 36.8	& 32.3 & 42.0
\\
\midrule
\textit{DiTTO} (DE) & 55.7	& \textbf{\cellcolor{green!40}48.2}	& 42.4	& 44.0	& \cellcolor{red!10}\textbf{36.5}	& 34.9 & 43.8 \\
\textit{DiTTO} (ES) & 55.5	& 42.5	& \textbf{\cellcolor{green!20}45.9}	& 45.7 & \cellcolor{red!10}\textbf{36.4}	& 34.8	& 43.5 \\
\textit{DiTTO} (FR) & 55.2	& 44.9	& 43.7	& \textbf{\cellcolor{green!40}46.4}	& \cellcolor{red!10}\textbf{36.5}	& 35.6	& 43.7
 \\
\textit{DiTTO} (JA) & 55.2	& \cellcolor{red!10}\textbf{40.9}	& \cellcolor{red!10}\textbf{41.3}	& \cellcolor{red!10}\textbf{42.5}	&  \textbf{\cellcolor{green!20}38.1}	& 33.9	& 42.0
 \\
\textit{DiTTO} (ZH) & \textbf{\cellcolor{green!40}55.8} & \cellcolor{red!10}\textbf{41.8}	& \cellcolor{red!10}\textbf{41.9}	& \cellcolor{red!10}\textbf{42.9}	& \cellcolor{red!10}\textbf{35.1}	& \textbf{\cellcolor{green!20}40.2} 	& 43.0
\\
\midrule
\textit{DiTTO} (UNF) & 55.0 & 46.4	& \textbf{\cellcolor{green!40}46.2}	& 45.9	& \textbf{\cellcolor{green!20}38.5}	& \cellcolor{green!20}40.4	& \textbf{\cellcolor{green!20}45.4}
\\
\DITTO & 55.3	& \textbf{\cellcolor{green!20}47.0}	& 45.1	& \textbf{\cellcolor{green!20}46.2}	& \textbf{\cellcolor{green!40}38.6}	& \textbf{\cellcolor{green!40}40.7}	& \textbf{\cellcolor{green!40}45.5}
\\
\bottomrule
\end{tabular}
\vspace{-0.5em}
\caption {Accuracy for single and  multiple target language variants of \DiTTO on MARC ($S$=1\%, mBERT).} 
\label{tbl:ditto_lang_sampling}
\vspace{-1em}
\end{table}

\hspace{-1em}\textbf{Target Language Prior Distribution}: In \DiTTO with multiple target language variant, the prior language distribution $p(\mathbb{T})$ is used to sample a target language for adaptation. To measure the importance of prior distribution, we experiment with two variants: (i) sampling based on the zero-shot performance of the Baseline method, which is used in the base setup of \DiTTO and (ii) \textit{DiTTO} (UNF) - with uniform sampling. Both the variants outperform Baseline with similar gains as shown in Table \ref{tbl:ditto_lang_sampling}. In the vanilla \DiTTO, where languages with lower zero-shot performance have a higher likelihood during sampling, provides better gains on these selected languages compared to the \textit{DiTTO} (UNF).

\hspace{-1em}\textbf{Task-Adaptive Pre-training (TAPT)}: The Baseline method does not utilize the available unlabeled data in the target languages, whereas \DiTTO uses the unlabeled data to improve the performance across all the target languages. Recently task-adaptive pre-training (TAPT) \cite{gururangan-etal-2020-dont, hossain-etal-2020-covidlies, caselli-etal-2021-hatebert} using unlabeled task-specific data has been shown to improve the performance for pre-trained language models across multiple tasks. However, TAPT has yet to be evaluated in a multilingual setting. 

To make a fair comparison, we have compared our proposed method with another baseline using unlabelled data, we shall refer this as \textit{Baseline (TAPT)}. \textit{TAPT} uses continued pre-training on the unlabeled target language data and fine-tuning is performed using the source language labelled dataset. We have reported the comparison between the new baseline method in Table \ref{tbl:new_baseline}. The \textit{Baseline (TAPT)} method outperforms the Baseline method where unlabeled data is not used in the source language ({\small EN}); however, it regresses for all the target languages. We hypothesize that the TAPT method generally improves the performance of the language used during fine-tuning. Still, it suffers from similar issues which the Baseline method suffers, such as low feature congruity in the fine-tuned features between the languages. \DiTTO, which does not suffer the feature incongruity issue, outperforms \textit{Baseline (TAPT)} for all the languages.

\begin{table}[ht]
\centering
\small 
\setlength\tabcolsep{3pt}%
\begin{tabular}{ccccccc} 
\toprule
Method &  {\small EN}  & {\small DE}  & {\small ES}      & {\small FR}  & {\small JA}  & {\small ZH}\\
\midrule 
Baseline & 56.40 & 56.02 & 53.29 & 52.15 & 49.77 & 48.05 
\\
\textit{Baseline (TAPT)} & \textbf{57.80} & {\color{red}55.58} & {\color{red}52.14} & {\color{red}51.70} & {\color{red}49.60} & {\color{red}45.71} \\
\DITTO & \textbf{61.28} & \textbf{59.06} & \textbf{54.87} & \textbf{55.50} & \textbf{53.29} & \textbf{51.01} 
\\
\bottomrule
\end{tabular}
\vspace{-0.5em}
\caption {Comparison of Baseline and \DiTTO methods with the new Baseline method using Task-Adaptive Pre-training (TAPT) on MARC ($S$=1\%, XLM-R).} 
\label{tbl:new_baseline}
\vspace{-1em}
\end{table}

\subsection{Congruity in Feature Representation}
\begin{figure}[!t]
\centering
\includegraphics[scale=0.35]{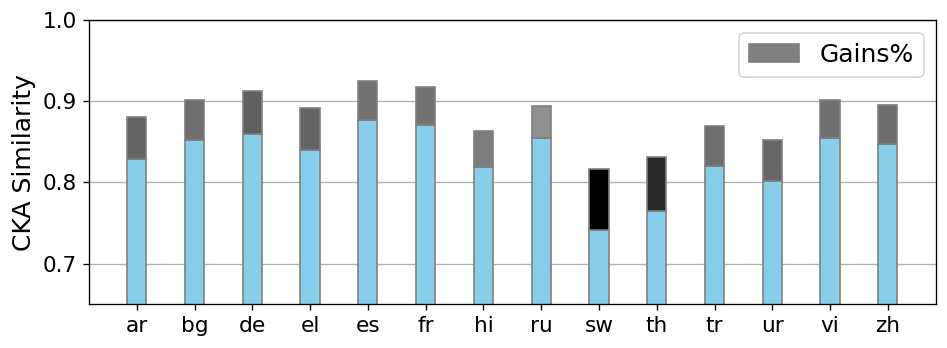}
\vspace{-0.5em}
\caption{Gains in CKA similarity (between features of source and target language) from \DiTTO over the Baseline method using mBERT on XNLI ($S$=10\%).
}
\label{fig:cka_ditto}
\vspace{-1em}
\end{figure}

As shown in Figure \ref{fig:baseline_cka} earlier that the zero-shot performance and feature congruity between the source and target languages are highly correlated. To validate our hypothesis that increasing the congruity between the features (via language adaptation) will improve the performance, we have plotted the increment in CKA similarity from \DiTTO over the Baseline method in Figure \ref{fig:cka_ditto}. We observe increment in CKA similarity across all the languages using \DiTTO, which is comparatively higher for distant or low-resource target languages. We also visualize the t-SNE projection of the feature representations of the source and the target languages in Figure \ref{fig:ditto_3dtsne}. It is difficult to distinguish between both languages in this figure, showcasing the quality of language adaptation.

\section{Discussion and Conclusion}
In this work, we propose a novel method to improve the cross-lingual transfer capability of pre-trained MMTs. We find that zero-shot performance is correlated with incongruency between the features of source and target languages. Experiments show that our proposed method \textit{DiTTO} outperforms the standard fine-tuning approach across multiple setups. In general, the gains from our method are higher on the models (as in mBERT) with less cross-lingual transfer. AmNLI consists only of languages that were not present in the pre-trained MMTs leading to similar transfer performance to the Baseline method. \DiTTO improves cross-lingual transfer using the pre-training features, hence the gains from \DiTTO are similar on both mBERT and XLM-R. Due to a similar reason, the relative gains for unseen languages do not follow the trend observed on seen languages, where the gains are higher for languages with the lower cross-lingual transfer. We find higher relative gains on unseen and low-resource languages, followed by distant languages. 
% The gains for seen languages are correlated with 
We also notice that the cross-lingual transfer improves with the amount of source language data $S$ for seen languages. In contrast, for unseen languages, improvements are limited. Due to this, the gains from \textit{DiTTO} start to decrease for high values of $S$ for seen languages but remain significant for unseen languages.

Our method provides similar gains using only unlabeled data compared to the fine-tuned Baseline  model (using 500 instances for each target language). Our ablation study shows that both LA and SAM are essential components of \textit{DiTTO}, with LA being the primary contributor to the gains. Experiments show that single language adaptation improves on that corresponding target language but may regress on other languages as the feature may remain incongruent to the source. However, \DITTO that adapts to multiple target languages performs best. \textit{DiTTO} tries to exploit the pre-training knowledge for improving the cross-lingual transfer, however few promising works such as adaptors \cite{pfeiffer-etal-2020-mad, ansell2021mad} have been proposed to improve the pre-training features for low-resource and unseen languages. However, task specific adaptors trained on the source language will also face the issue of incongruity in the feature representations. Hence, adaptors will not improve the cross-lingual transfer, but only improves the pre-trained features. We plan to extend our method towards integrating with adaptors to take advantage of pre-training features and improve performance.

\section{Limitations}
Unlabeled data in the target language is essential for the proposed method \DiTTO for improving cross-lingual transfer. Obtaining unlabeled data can be challenging for specific tasks where the proposed approach may not be applicable. However, we recommend using the \textit{DiTTO}-LA variant for these scenarios. Another limitation of \DiTTO is that it requires all the target languages to be present during the fine-tuning stage to obtain the performances mentioned in our work, which might not be viable for all the tasks. Nevertheless, the gains from \DiTTO may transfer to the new target languages if these languages are typologically similar to the target languages used during the fine-tuning of \DiTTO. In the vanilla setup of \DiTTO, the prior language probability depends upon the zero-shot accuracy using the Baseline method, which requires a validation or test dataset in each target language. This dependency may limit its application. However, \textit{DiTTO} (UNF) can be used for obtaining similar gains if the validation sets are not available.

\bibliographystyle{acl_natbib}
\bibliography{anthology,custom}

\newpage
\appendix

\section{Data Statistics}
We have provided the statistics of training and test data after removing any duplicates in each of the target languages for all the datasets in Tables \ref{tbl:xnli_dataset}, \ref{tbl:amnli_dataset}, and \ref{tbl:marc_dataset}. 

\begin{table*}[ht]
\centering
\setlength\tabcolsep{5pt}%
\begin{tabular}{*{200}{c}}
\toprule
\textbf{ISO} & \textbf{Language} & Train & Test & XLM-R Group & mBERT Group\\ 
  \midrule
{\small AR} & Arabic & 392403 & 5010 & Distant & Distant \\
{\small BG} & Bulgarian & 392335 & 5010 & Distant & Distant \\
{\small DE} & German & 392440 & 5010 & Similar & Similar \\
{\small EL} & Greek & 392331 & 5010 & Distant & Distant \\
{\small EN} & English & 392568 & 5010  & Source & Source \\
{\small ES} & Spanish & 392405 & 5010 & Similar & Similar \\
{\small FR} & French & 392405 & 5010 & Similar & Similar \\
{\small HI} & Hindi & 392356 & 5010  & Distant & Low-Resource \\
{\small RU} & Russian & 392318 & 5010  & Similar & Similar \\
{\small SW} & Swahili & 391819 & 5010  & Low-Resource & Low-Resource \\
{\small TH} & Thai & 392480 & 5010 & Distant & Low-Resource  \\
{\small TR} & Turkish & 392177 & 5010 & Distant & Distant \\
{\small UR} & Urdu & 388826 & 5010 & Low-Resource & Low-Resource  \\
{\small VI} & Vietnamese & 392416 & 5010 & Distant & Distant \\
{\small ZH} & Chinese & 392251 & 5010 & Distant & Distant \\
 \bottomrule 
\end{tabular}
\caption {In this table, we have reported the target language categories and statistics of training and test data available in each language for XNLI dataset.}
  \label{tbl:xnli_dataset}
 \end{table*}
 
 \begin{table*}[ht]
\centering
\setlength\tabcolsep{5pt}%
\begin{tabular}{*{200}{c}}
\toprule
\textbf{ISO} & \textbf{Language} & Train & Test & XLM-R Group & mBERT Group\\ 
  \midrule
{\small AYM} & Aymara & 743 & 750 & Unseen & Unseen\\
{\small CNI} & Asháninka & 658 & 750 & Unseen & Unseen\\
{\small BZD} & Bribri & 743 & 750 & Unseen & Unseen\\
{\small GN} & Guaraní & 743 & 750 & Unseen & Unseen\\
{\small NAH} & Nahuatl & 376 & 738 & Unseen & Unseen\\
{\small OTO} & Otomí & 222 & 748 & Unseen & Unseen\\
{\small QUY} & Quechua & 743 & 750 & Unseen & Unseen\\
{\small TAR} & Rarámuri & 743 & 750 & Unseen & Unseen\\
{\small SHP} & Shipibo-Konibo & 743 & 750 & Unseen & Unseen \\
{\small HCH} & Wixarika & 743 & 750 & Unseen & Unseen \\
 \bottomrule 
\end{tabular}
\caption {In this table, we have reported the target language categories and statistics of training and test data available in each language for AmNLI dataset.}
  \label{tbl:amnli_dataset}
 \end{table*}

 \begin{table*}[ht]
\centering
\setlength\tabcolsep{5pt}%
\begin{tabular}{*{200}{c}}
\toprule
\textbf{ISO} & \textbf{Language} & Train & Test & XLM-R Group & mBERT Group\\ 
  \midrule
{\small DE} & German & 199877 & 4993 & Similar & Similar\\
{\small EN} & English & 199891 & 4998 & Source & Source\\
{\small ES} & Spanish & 199726 & 4986 & Similar & Similar\\
{\small FR} & French & 199612 & 4986 & Similar & Similar\\
{\small JA} & Japanese & 199845 & 4995 & Distant & Distant\\
{\small ZH} & Chinese & 197418 & 4903 & Distant & Distant\\
 \bottomrule 
\end{tabular}
\caption {In this table, we have reported the target language categories and statistics of training and test data available in each language for MARC dataset.}
  \label{tbl:marc_dataset}
 \end{table*}
 
% \section{Chinese as Source Language}
%  To measure the effectiveness of \DiTTO across different source languages, we conduct zero-shot experiments considering Chinese ({\small ZH}) as the source language in the MARC dataset. We have reported the average accuracy across all the languages in Table \ref{tbl:relative_gain_ditto_zh}. \DiTTO provides consistent gains over the Baseline method across all the training configurations, comparatively higher gains than {\small EN} as the source language. 

% \begin{table*}[ht]
% \centering
% % \setlength\tabcolsep{4pt}%
% \begin{tabular}{c|ccc|ccc}
% \toprule
% \multicolumn{1}{c}{} & \multicolumn{3}{c}{mBERT}  & \multicolumn{3}{|c}{XLM-R} \\
%     \cmidrule(lr){2-4}\cmidrule(lr){5-7}
%  \multirow{-1}{*}{\textbf{Dataset}} & 1\%  & 10\%&  100\% & 1\%  & 10\%&  100\% \\ 
% \midrule
% Baseline & 32.88 & 39.48 & 42.68 & 45.83 & 50.86 & 51.38 \\
% \DITTO & \textbf{39.09} & \textbf{46.90} & \textbf{50.82} & \textbf{51.84} & \textbf{53.34} & \textbf{55.27} \\
% \midrule
% RG(\%) & \textbf{19.45} & \textbf{20.80} & \textbf{20.67} & \textbf{13.31} & 5.34 & 8.12 \\
% \bottomrule
% \end{tabular}
% \vspace{-0.5em}
% \caption {We have reported the zero-shot accuracy averaged across all languages with \textbf{{\small ZH}} as the source language data on MARC dataset. RG denotes the relative gains averaged across all the languages from using \DITTO over Baseline.} 
% \label{tbl:relative_gain_ditto_zh}
% \end{table*}

\section{Performance and Cost Trade-off}
\label{section:cost}
\begin{figure*}[!h]
\centering
\includegraphics[scale=0.4]{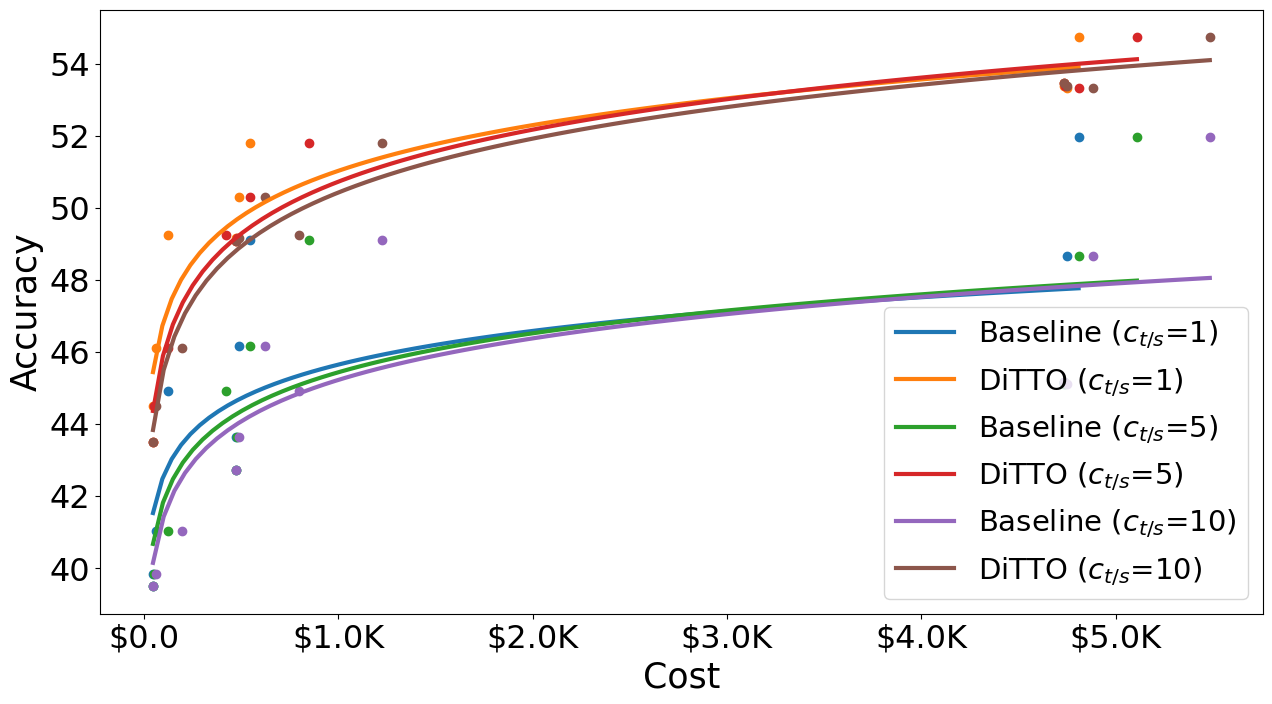}
\vspace{-0.7em}
\caption{The plot shows Accuracy (vs) Cost graph with various values of $c_{t/s}$ for \DITTO and Baseline method trained using mBERT on XNLI ($S$=10\%) dataset.
        }
\label{fig:cost_performance_mbert}
\vspace{-0.5em}
\end{figure*}

\begin{figure*}[!ht]
\centering
\includegraphics[scale=0.4]{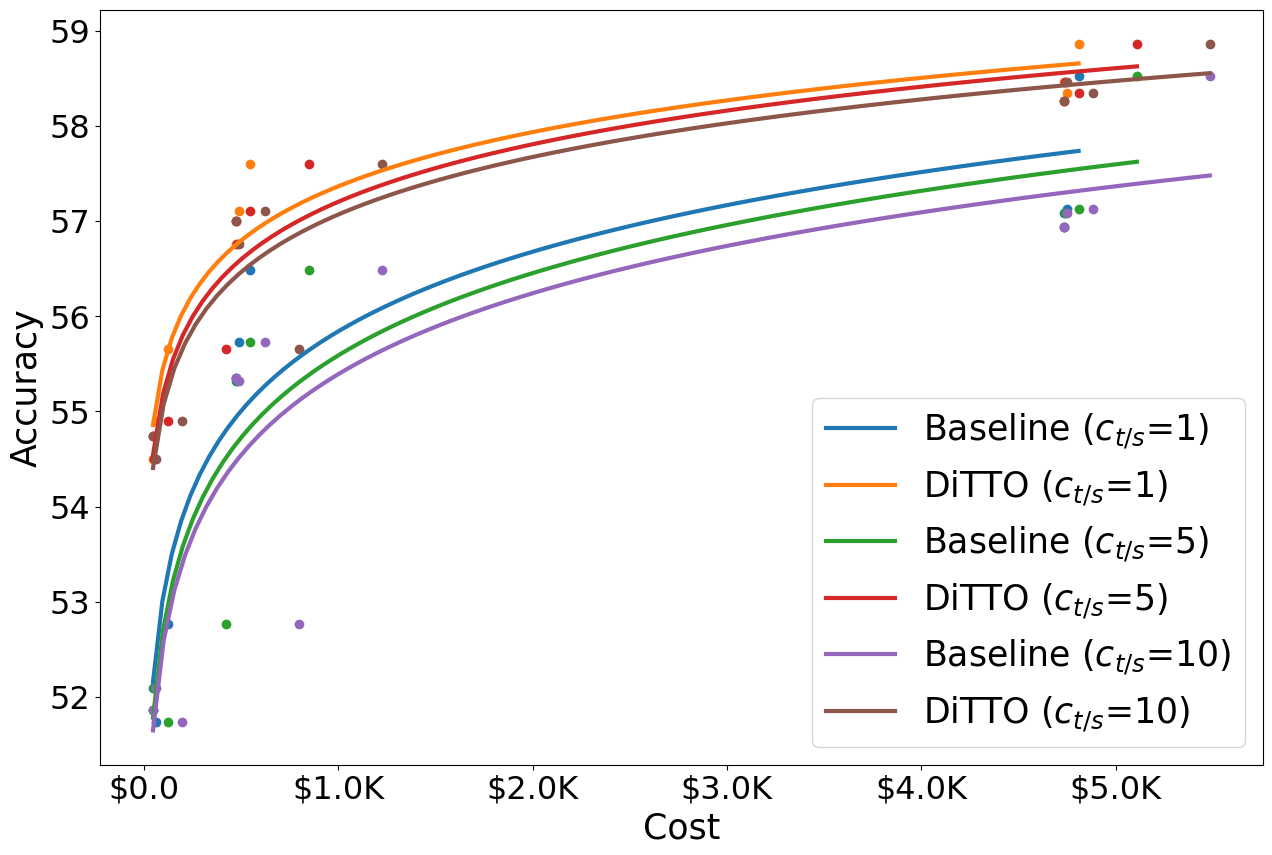}
\caption{The plot shows Accuracy (vs) Cost graph with various values of $c_{t/s}$ for \DITTO and Baseline method trained using XLM-R on XNLI ($S$=10\%) dataset.}
\label{fig:cost_performance_xlmr}
\end{figure*}

From the above results, it seems that \DiTTO is more cost-effective in terms of both source and target language data
% \footnote{Across various values of source language data ($S$) and few-shot target language data ($k$)}. 
We validate this by plotting the accuracy from both methods against the cost incurred while collecting the labeled data for fine-tuning. Assuming there is no cost associated with collecting unlabeled data, we define the cost $\mathbb{C}$ for building a fine-tuning dataset as follows:
\vspace{-0.5em}
\begin{gather}
    \mathbb{C} = c_s * n^l_s + c_s * c_{t/s} * k * |\mathbb{T}|
\end{gather}

where $c_s$ is the cost of obtaining one instance labeled in the source language and we assume it to be 3 cents considering {\small EN} as the source language. $c_{t/s}$ is the relative cost of obtaining labeled data in target language compared to the source language. We use Gaussian Process Regression with a dot product kernel for modeling performance with cost. In Figure \ref{fig:cost_performance_mbert} and \ref{fig:cost_performance_xlmr}, we plot the accuracy for various values of $c_{t/s}$ against the total cost incurred using mBERT on the MARC dataset, we observe a convex curve with increasing curvature as the value of $c_{t/s}$ increases. From the plot, we can see that higher accuracy can be achieved using \DiTTO than Baseline at the same cost for all the values of $c_{t/s}$, showing the cost-saving nature of \DiTTO with average savings of 7 times.

% \section{Calibration Errors}

% \begin{figure}[!ht]
% \centering
% \includegraphics[scale=0.35]{figures/fewshot_source_data_mbert.png}
% \caption{The plot shows CKA similarity
% between the features of source language (EN ) and the target languages using \DITTO and Baseline method trained using mBERT on XNLI-10 dataset.
% }
% \label{fig:fewshot_source_data_mbert}
% \end{figure}

% \begin{table*}[ht]
% \centering
% % \setlength\tabcolsep{3pt}%
% \begin{tabular}{cc|ccc|ccc}
% \toprule
% \multicolumn{1}{c}{} & \multicolumn{1}{c}{} & \multicolumn{3}{c}{mBERT}  & \multicolumn{3}{|c}{XLM-R} \\
%     \cmidrule(lr){3-5}\cmidrule(lr){6-8}
%  \multirow{-1}{*}{\textbf{Dataset}} & \multirow{-1}{*}{\textbf{|$\mathbb{T}$|}} & 1\%  & 10\%&  100\% &1\%  & 10\%&  100\% \\ 
% \midrule
% XNLI & 14 & 10.3 & 12.3 & 15.5 & 8.3 & 10.3 & 11.3 \\
% AmNLI & 10 & {\color{red} 24.8} & {\color{red} 32.9} & {\color{red} {41.2}} & {\color{red} 29.9} & {\color{red} 39.2} & {\color{red} {45.1}} \\
% MARC & 5 & 14.8 & 18.1 & 20.3 & 4.5 & 8.8 & 9.9  \\
% % PAWS-X & 6 & 9.9 & 12.9 & 12.0 & 4.2 & 10.9 & 11.6 \\
% \bottomrule
% \end{tabular}
% \vspace{-0.5em}
% \caption {We have reported the mean of difference $\triangle$ between the zero-shot performance of all the target languages and source language for varying amount of the source language ({\small EN}) data used while fine-tuning. |$\mathbb{T}$| is the number target languages available in the dataset.} 
% \label{tbl:zeroshot_delta_appendix}
% \end{table*}

\begin{table*}[!h]
\centering
\setlength\tabcolsep{3pt}%
\begin{tabular}{c|ccc|ccc|ccc}
\toprule
\multicolumn{1}{c}{} & \multicolumn{3}{c}{$S=1\%$}  & \multicolumn{3}{|c}{$S=10\%$} & \multicolumn{3}{|c}{$S=100\%$}\\
    \cmidrule(lr){2-5}\cmidrule(lr){5-7} \cmidrule(lr){8-10}
 \multicolumn{1}{c|}{Language} & Baseline  & \DITTO  & RG & Baseline  & \DITTO  & RG & Baseline  & \DITTO  & RG \\
\midrule
\multicolumn{10}{c}{XLM-R} \\
\midrule
EN	& 67.84	& 69.00	& 1.68	& 78.08	& 79.24	& 1.48	& 83.83	& 82.59	& -1.48\\
AYM	& 37.47	& 43.60	& 16.37	& 36.00	& 46.93	& 30.37	& 36.27	& 45.73	& 26.10\\
BZD	& 35.73	& 50.40	& 41.04	& 38.13	& 54.40	& 42.66	& 38.40	& 55.07	& 43.40\\
CNI	& 37.60	& 44.27	& 17.73	& 38.13	& 42.80	& 12.24	& 39.07	& 42.27	& 8.19\\
GN	& 42.46	& 49.53	& 16.67	& 42.86	& 56.21	& 31.15	& 39.92	& 48.60	& 21.74\\
HCH	& 33.51	& 41.92	& 25.10	& 38.72	& 41.39	& 6.90	& 37.92	& 41.39	& 9.15\\
NAH	& 38.89	& 44.99	& 15.68	& 42.01	& 45.26	& 7.74	& 42.14	& 46.21	& 9.65\\
OTO	& 37.43	& 42.91	& 14.64	& 38.24	& 43.72	& 14.34	& 38.90	& 45.45	& 16.84\\
QUY	& 37.47	& 42.27	& 12.81	& 37.60	& 43.87	& 16.67	& 37.20	& 46.53	& 25.09\\
SHP	& 38.93	& 46.13	& 18.49	& 42.27	& 46.67	& 10.41	& 41.07	& 45.73	& 11.36\\
TAR	& 40.05	& 40.45	& 1.00	& 35.11	& 44.33	& 26.24	& 36.45	& 43.52	& 19.41\\
\midrule
AVG	& 37.95	& 44.65	& 17.95	& 38.91	& 46.56	& 19.87	& 38.73	& 46.05	& 19.09\\
\midrule
\multicolumn{10}{c}{mBERT} \\
\midrule
EN	& 62.53	& 64.83	& 3.67	& 71.20	& 73.05	& 2.61	& 81.18	& 79.64	& -1.89\\
AYM	& 38.27	& 44.93	& 17.42	& 38.93	& 47.07	& 20.89	& 39.33	& 47.07	& 19.66\\
BZD	& 34.80	& 44.00	& 26.44	& 37.47	& 45.60	& 21.71	& 42.13	& 45.60	& 8.23\\
CNI	& 37.60	& 39.87	& 6.03	& 37.47	& 47.47	& 26.69	& 40.00	& 44.93	& 12.33\\
GN	& 40.19	& 46.86	& 16.61	& 38.85	& 49.80	& 28.18	& 41.79	& 51.67	& 23.64\\
HCH	& 34.98	& 40.85	& 16.79	& 36.98	& 45.79	& 23.83	& 39.92	& 44.59	& 11.71\\
NAH	& 40.79	& 44.72	& 9.63	& 42.28	& 46.07	& 8.97	& 43.90	& 48.92	& 11.42\\
OTO	& 38.10	& 38.64	& 1.40	& 37.43	& 41.58	& 11.07	& 37.97	& 44.39	& 16.90\\
QUY	& 38.67	& 39.47	& 2.07	& 36.53	& 42.80	& 17.15	& 38.00	& 43.87	& 15.44\\
SHP	& 38.40	& 40.53	& 5.56	& 40.13	& 46.93	& 16.94	& 41.73	& 46.67	& 11.82\\
TAR	& 35.91	& 40.99	& 14.13	& 36.85	& 44.86	& 21.74	& 35.65	& 42.72	& 19.85\\
\midrule
AVG	& 37.77	& 42.09	& 11.61	& 38.29	& 45.80	& 19.72	& 40.04	& 46.04	& 15.10\\
\bottomrule
\end{tabular}
\vspace{-0.5em}
\caption {We have reported the accuracy and relative gains using XLM-R and mBERT on AmNLI dataset. The average relative gain is denotes the average gains across all the languages except the source {\small EN}.} 
\label{tbl:amnli_lang}
\end{table*}

\begin{table*}[!h]
\centering
\setlength\tabcolsep{3pt}%
\begin{tabular}{c|ccc|ccc|ccc}
\toprule
\multicolumn{1}{c}{} & \multicolumn{3}{c}{$S=1\%$}  & \multicolumn{3}{|c}{$S=10\%$} & \multicolumn{3}{|c}{$S=100\%$}\\
    \cmidrule(lr){2-5}\cmidrule(lr){5-7} \cmidrule(lr){8-10}
 \multicolumn{1}{c|}{Language} & Baseline  & \DITTO  & RG & Baseline  & \DITTO  & RG & Baseline  & \DITTO  & RG \\
\midrule
\multicolumn{10}{c}{XLM-R} \\
\midrule
AR	& 55.25	& 65.19	& 17.99	& 64.69	& 67.96	& 5.06	& 71.16	& 73.39	& 3.14\\
BG	& 60.78	& 68.84	& 13.27	& 70.56	& 73.87	& 4.70	& 76.59	& 78.16	& 2.06\\
DE	& 60.98	& 67.50	& 10.70	& 70.64	& 71.92	& 1.81	& 75.33	& 77.37	& 2.70\\
EL	& 59.78	& 66.75	& 11.65	& 68.72	& 71.02	& 3.34	& 74.91	& 76.47	& 2.08\\
EN	& 66.49	& 72.91	& 9.67	& 77.80	& 79.72	& 2.46	& 83.71	& 84.65	& 1.12\\
ES	& 63.77	& 69.44	& 8.89	& 72.46	& 74.99	& 3.50	& 77.17	& 79.12	& 2.53\\
FR	& 62.51	& 68.86	& 10.15	& 71.52	& 73.77	& 3.15	& 76.85	& 78.50	& 2.16\\
HI	& 54.85	& 63.81	& 16.34	& 64.07	& 67.05	& 4.64	& 68.98	& 71.32	& 3.39\\
RU	& 60.32	& 66.67	& 10.52	& 69.62	& 71.66	& 2.92	& 74.49	& 76.93	& 3.27\\
SW	& 51.86	& 60.16	& 16.01	& 61.34	& 63.75	& 3.94	& 65.67	& 66.39	& 1.09\\
TH	& 55.81	& 64.75	& 16.02	& 63.89	& 68.14	& 6.65	& 70.96	& 73.45	& 3.52\\
TR	& 57.88	& 65.89	& 13.83	& 67.62	& 69.98	& 3.48	& 71.82	& 74.03	& 3.09\\
UR	& 53.17	& 62.12	& 16.82	& 61.94	& 65.91	& 6.41	& 64.83	& 66.95	& 3.26\\
VI	& 58.56	& 66.65	& 13.80	& 68.82	& 72.12	& 4.79	& 74.05	& 75.99	& 2.61\\
ZH	& 58.58	& 66.79	& 14.00	& 68.46	& 70.52	& 3.00	& 73.25	& 75.43	& 2.97\\
\midrule
Average	& 58.15	& 65.96	& 13.57	& 67.45	& 70.19	& 4.10	& 72.57	& 74.54	& 2.71\\
\midrule
\multicolumn{10}{c}{mBERT} \\
\midrule
AR	& 47.09	& 56.75	& 20.52	& 57.78	& 62.87	& 8.81	& 63.07	& 65.21	& 3.39\\
BG	& 50.00	& 60.26	& 20.52	& 63.31	& 66.47	& 4.98	& 68.78	& 68.50	& -0.41\\
DE	& 49.44	& 60.10	& 21.56	& 65.35	& 67.92	& 3.94	& 70.00	& 72.02	& 2.88\\
EL	& 48.70	& 59.08	& 21.31	& 60.12	& 64.63	& 7.50	& 65.91	& 66.99	& 1.64\\
EN	& 57.17	& 64.87	& 13.48	& 72.00	& 74.97	& 4.13	& 81.34	& 82.67	& 1.64\\
ES	& 50.12	& 62.38	& 24.45	& 66.11	& 70.88	& 7.22	& 73.11	& 75.43	& 3.17\\
FR	& 51.96	& 61.40	& 18.17	& 67.52	& 69.06	& 2.28	& 72.63	& 74.91	& 3.13\\
HI	& 46.57	& 54.93	& 17.96	& 57.25	& 60.58	& 5.82	& 60.26	& 62.02	& 2.91\\
RU	& 49.64	& 58.82	& 18.50	& 63.39	& 66.35	& 4.66	& 67.70	& 68.98	& 1.89\\
SW	& 37.82	& 46.51	& 22.96	& 45.91	& 49.20	& 7.17	& 50.68	& 49.42	& -2.48\\
TH	& 36.61	& 53.31	& 45.64	& 48.20	& 56.21	& 16.60	& 53.85	& 57.03	& 5.89\\
TR	& 45.35	& 57.25	& 26.23	& 58.22	& 61.26	& 5.21	& 62.20	& 61.42	& -1.25\\
UR	& 45.19	& 53.91	& 19.30	& 54.83	& 59.10	& 7.79	& 58.74	& 59.64	& 1.53\\
VI	& 49.20	& 59.98	& 21.91	& 63.45	& 67.25	& 5.98	& 69.46	& 71.44	& 2.84\\
ZH	& 48.74	& 60.26	& 23.63	& 63.97	& 66.63	& 4.15	& 68.64	& 71.60	& 4.30\\
\midrule
Average	& 46.89	& 57.50	& 23.05	& 59.67	& 63.46	& 6.58	& 64.65	& 66.04	& 2.10\\
\bottomrule
\end{tabular}
\vspace{-0.5em}
\caption {We have reported the accuracy and relative gains  using XLM-R and mBERT on XNLI dataset. The average relative gain is denotes the average gains across all the languages except the source {\small EN}.} 
\label{tbl:xnli_lang}
\end{table*}

\begin{table*}[!h]
\centering
\setlength\tabcolsep{3pt}%
\begin{tabular}{c|ccc|ccc|ccc}
\toprule
\multicolumn{1}{c}{} & \multicolumn{3}{c}{$S=1\%$}  & \multicolumn{3}{|c}{$S=10\%$} & \multicolumn{3}{|c}{$S=100\%$}\\
    \cmidrule(lr){2-5}\cmidrule(lr){5-7} \cmidrule(lr){8-10}
 \multicolumn{1}{c|}{Language} & Baseline  & \DITTO  & RG & Baseline  & \DITTO  & RG & Baseline  & \DITTO  & RG \\
 \midrule
\multicolumn{10}{c}{XLM-R} \\
\midrule
 EN	& 56.40	& 61.28	& 8.66	& 64.17	& 64.37	& 0.31	& 66.81	& 66.91	& 0.15\\
DE	& 56.02	& 59.06	& 5.43	& 60.54	& 62.11	& 2.58	& 63.31	& 64.11	& 1.27\\
ES	& 53.29	& 54.87	& 2.97	& 56.34	& 56.96	& 1.10	& 57.68	& 58.80	& 1.95\\
FR	& 52.15	& 55.50	& 6.42	& 56.62	& 57.72	& 1.95	& 58.44	& 58.76	& 0.55\\
JA	& 49.77	& 53.29	& 7.08	& 52.93	& 55.46	& 4.77	& 53.77	& 55.98	& 4.10\\
ZH	& 48.05	& 51.01	& 6.15	& 50.34	& 52.74	& 4.78	& 51.50	& 53.66	& 4.20\\
\midrule
Average	& 51.86	& 54.75	& 5.61	& 55.35	& 57.00	& 3.04	& 56.94	& 58.26	& 2.41\\
\midrule
\multicolumn{10}{c}{mBERT} \\
\midrule
EN	& 54.32	& 55.82	& 2.76	& 60.80	& 62.77	& 3.22	& 65.53	& 65.71	& 0.27\\
DE	& 42.32	& 46.75	& 10.46	& 44.30	& 52.55	& 18.63	& 48.61	& 58.90	& 21.18\\
ES	& 42.30	& 45.81	& 8.30	& 45.77	& 51.18	& 11.83	& 49.56	& 54.75	& 10.48\\
FR	& 43.76	& 47.31	& 8.11	& 48.28	& 51.42	& 6.52	& 49.74	& 55.31	& 11.21\\
JA	& 36.82	& 38.66	& 5.00	& 39.00	& 43.78	& 12.27	& 39.32	& 48.77	& 24.03\\
ZH	& 32.31	& 41.85	& 29.55	& 36.32	& 46.40	& 27.74	& 38.67	& 49.60	& 28.27\\
\midrule
Average	& 39.50	& 44.08	& 12.28	& 42.73	& 49.07	& 15.40	& 45.18	& 53.47	& 19.03\\
\bottomrule
\end{tabular}
\vspace{-0.5em}
\caption {We have reported the accuracy and relative gains using XLM-R and mBERT on MARC dataset. The average relative gain is denotes the average gains across all the languages except the source {\small EN}.} 
\label{tbl:marc_lang}
\end{table*}

\begin{figure*}[!ht]
    \centering
    \begin{subfigure}[t]{\textwidth}
        \centering
\includegraphics[scale=0.3]{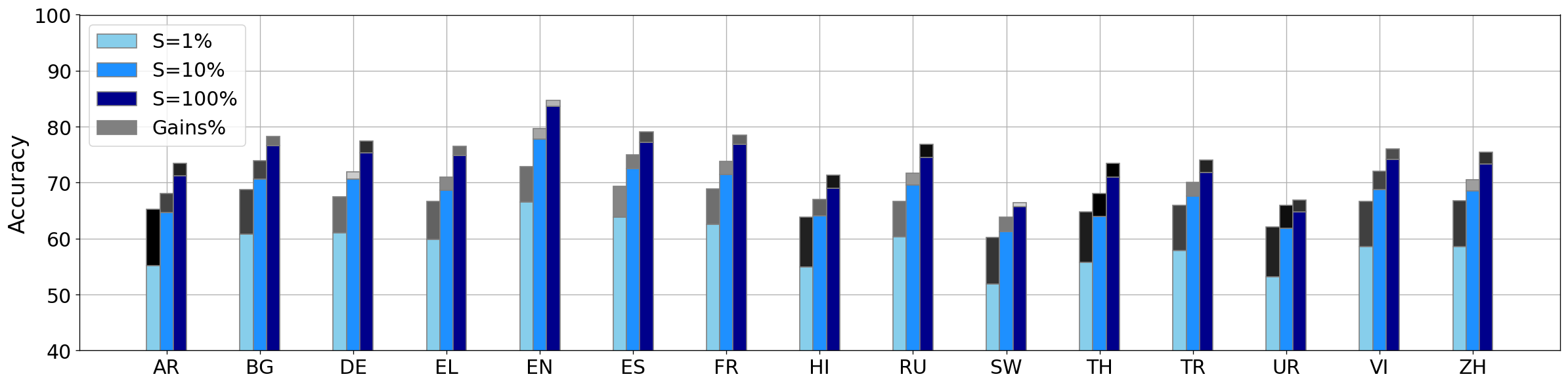}
         \caption{XLM-R}
    \end{subfigure}
    \begin{subfigure}[t]{\textwidth}
        \centering
\includegraphics[scale=0.3]{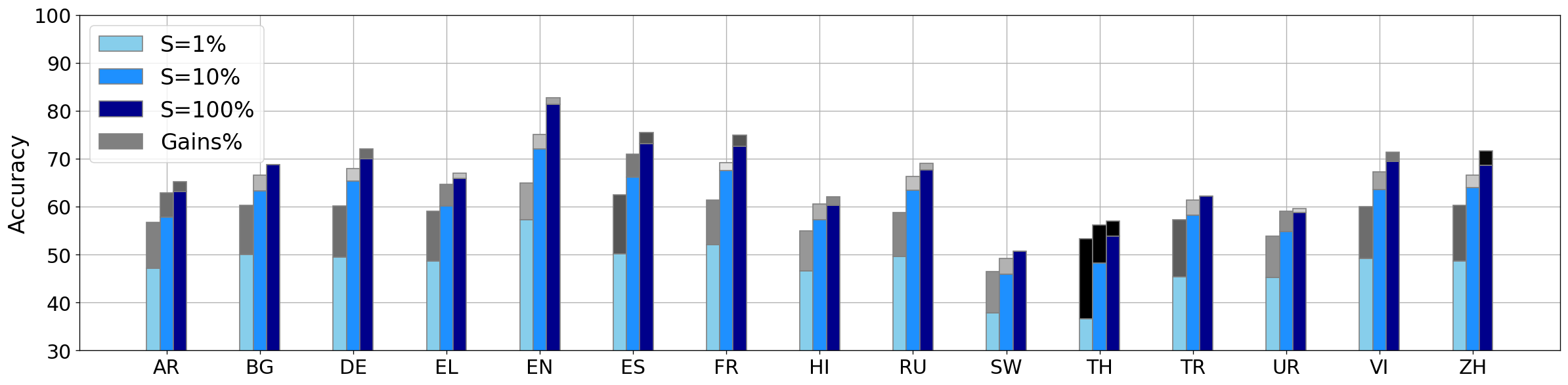}
        \caption{mBERT}
    \end{subfigure}
\caption{Absolute gains (darker shades of grey denotes higher gains) from \DITTO across all target languages on XNLI dataset.}
\label{fig:language_xnli}
\end{figure*}

\begin{figure*}[!ht]
    \centering
    \begin{subfigure}[t]{\textwidth}
        \centering
\includegraphics[scale=0.5]{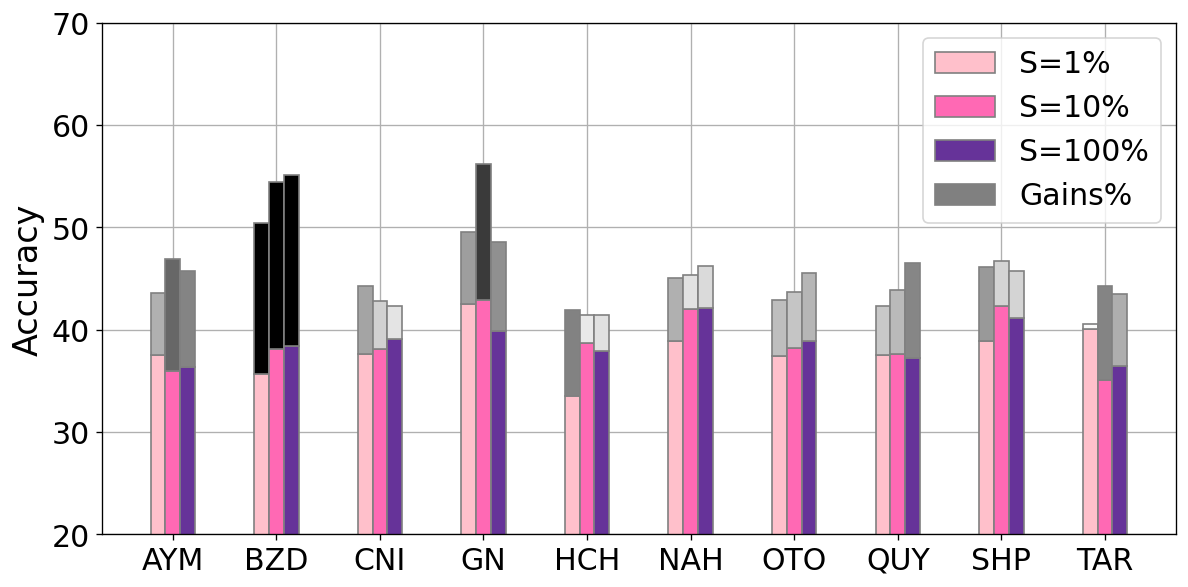}
         \caption{XLM-R}
    \end{subfigure}
    \begin{subfigure}[t]{\textwidth}
        \centering
\includegraphics[scale=0.5]{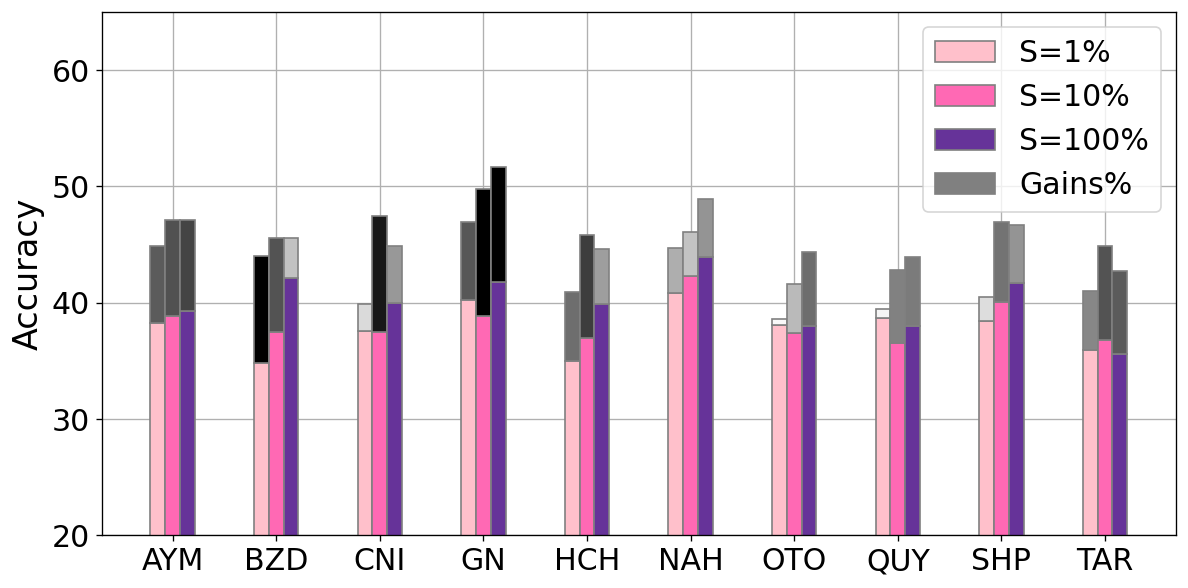}
        \caption{mBERT}
    \end{subfigure}
\caption{Absolute gains (darker shades of denotes higher gains) from \DITTO across all target languages on AmNLI dataset.}
    \label{fig:language_amli}
\end{figure*}

\begin{figure*}[!ht]
    \centering
    \begin{subfigure}[t]{\textwidth}
        \centering
\includegraphics[scale=0.5]{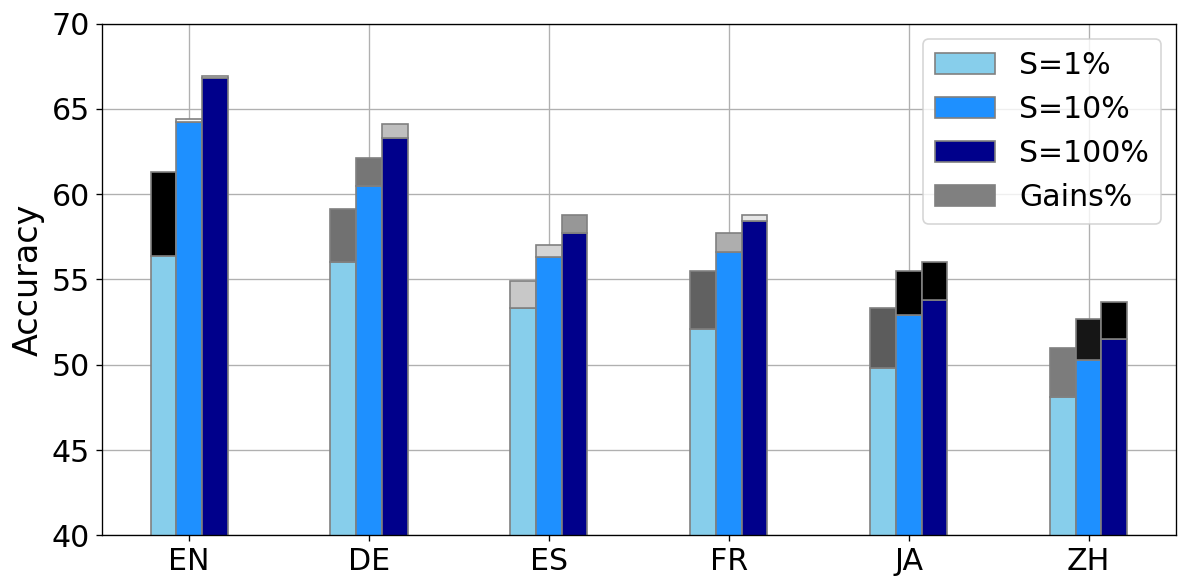}
         \caption{XLM-R}
    \end{subfigure}
    \begin{subfigure}[t]{\textwidth}
        \centering
\includegraphics[scale=0.5]{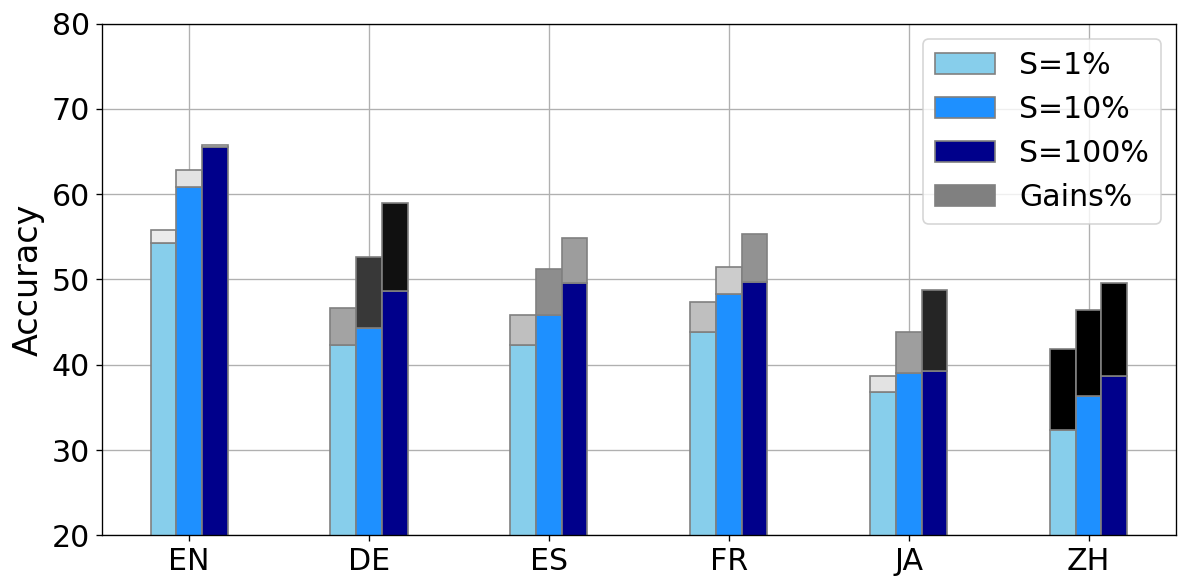}
        \caption{mBERT}
    \end{subfigure}
\caption{Absolute gains (darker shades of denotes higher gains) from \DITTO across all target languages on MARC dataset.}
    \label{fig:language_marc}
\end{figure*}

\begin{figure*}[!ht]
\centering
\includegraphics[scale=0.4]{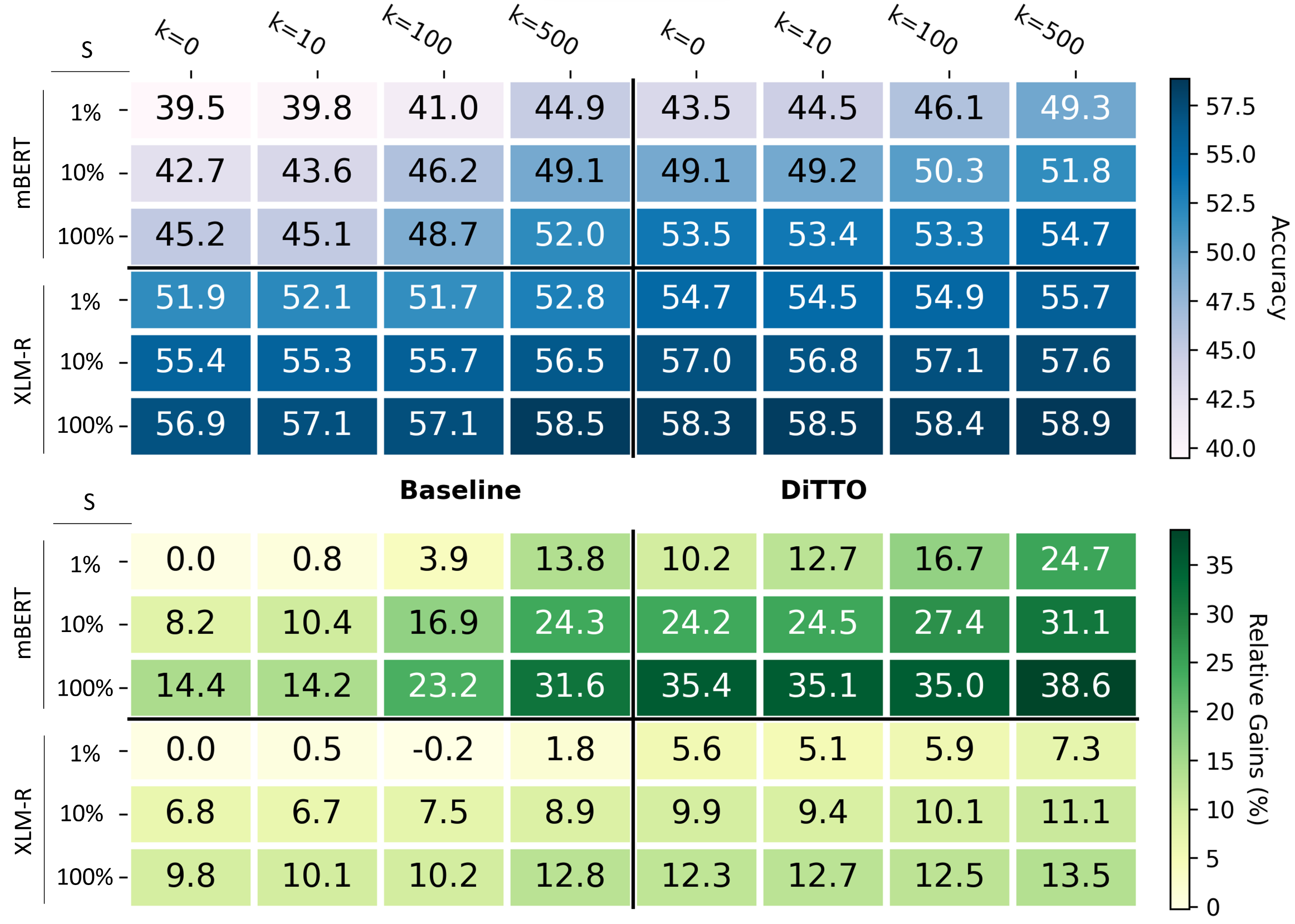}
\caption{Accuracy/relative gains\footnote{The relative gain is calculated with respect to the accuracy of the Baseline method on $S=1\%$ and $k=0$.} on MARC dataset. Rows and columns denoting the amount of source and target language labeled instances, respectively.
}
\label{fig:heatmap_fewshot_marc_v2}
\end{figure*}

\begin{figure*}[!ht]
\centering
\includegraphics[scale=0.4]{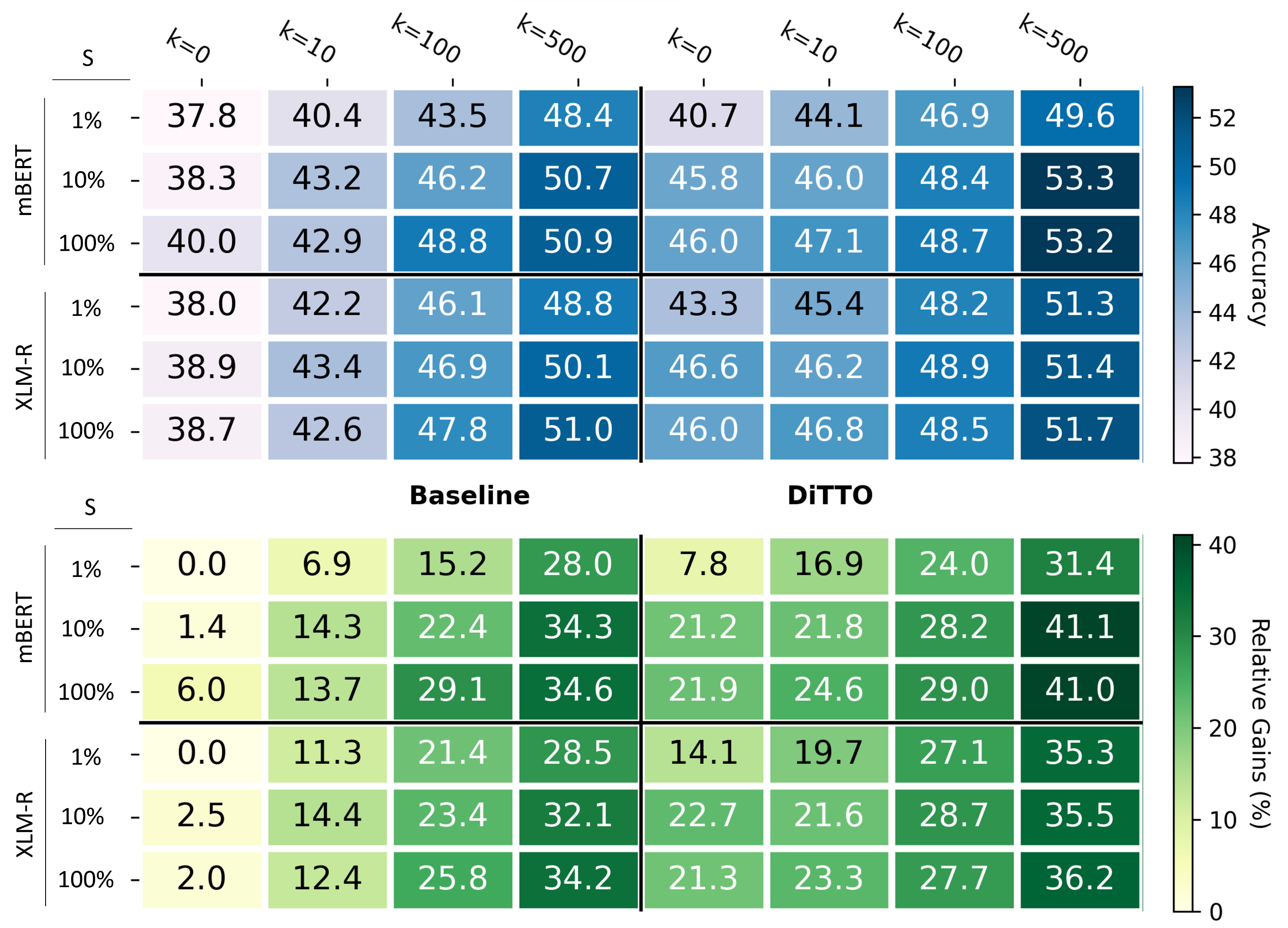}
\caption{Accuracy/relative gains\footnote{The relative gain is calculated with respect to the accuracy of the Baseline method on $S=1\%$ and $k=0$.} on AmNLI dataset. Rows and columns denoting the amount of source and target language labeled instances, respectively.
}
\label{fig:heatmap_fewshot_amnli}
\end{figure*}

% \begin{table*}[!h]
% \centering
% \setlength\tabcolsep{3pt}%
% \begin{tabular}{c|c|ccc|ccc}
% \toprule
% \multicolumn{1}{c}{} &  &\multicolumn{3}{c}{mBERT}  & \multicolumn{3}{|c}{XLM-R} \\
%     \cmidrule(lr){3-5}\cmidrule(lr){6-8}
%  \multicolumn{1}{c|}{Dataset} &\multirow{-1}{*}{\textbf{Method}} & 1\%  & 10\%&  100\% & 1\%  & 10\%&  100\% \\ 
% \midrule
% & \textit{Baseline} & 0.27	& \textbf{0.07}	& 0.11	& 0.25	& 0.11	& \textbf{0.06}\\
% \multirow{-3}{*}{XNLI} &  \DITTO & \textbf{0.26}	& 0.08	& \textbf{0.11}	& \textbf{0.17}	& \textbf{0.08}	& 0.07\\
% \midrule
%  & \textit{Baseline} & 0.16	& \textbf{0.07}	& \textbf{0.04}	& \textbf{0.04}	& 0.05	& \textbf{0.03}\\
% \multirow{-3}{*}{MARC} &  \DITTO & \textbf{0.11}	& \textbf{0.07}	& \textbf{0.04}	& 0.05	& \textbf{0.04}	& \textbf{0.03}\\
% \midrule
% & \textit{Baseline} & 0.37	& 0.32	& 0.24	& 0.36	& 0.32	& 0.20\\
% \multirow{-3}{*}{AmNLI} &  \DITTO & \textbf{0.33}	& \textbf{0.09}	& \textbf{0.17}	& \textbf{0.29}	& \textbf{0.21}	& \textbf{0.19}\\
% \bottomrule
% \end{tabular}
% \vspace{-0.5em}
% \caption {We have reported the calibration errors averaged across all the languages.} 
% \label{tbl:calibration}
% \end{table*}

\end{document}